\documentclass[]{fairmeta}
\usepackage{wrapfig}    % Added for wrapfigure
\usepackage{latexsym}
\usepackage{microtype}
\usepackage{hyperref}
\usepackage{url}
\usepackage{booktabs}
\usepackage[toc,page]{appendix}
\usepackage{titletoc}
\usepackage{lineno}
\usepackage[normalem]{ulem}
\useunder{\uline}{\ul}{}
\usepackage[export]{adjustbox}
\usepackage{graphicx}
\usepackage{rotating}
\usepackage{subcaption}

\usepackage{xcolor} % For text coloring
\usepackage{colortbl} % For table cell coloring 
\usepackage{tcolorbox} % Required for creating colored boxes

\definecolor{darkblue}{rgb}{0, 0, 0.5}
\usepackage[dvipsnames]{xcolor}

\usepackage{graphicx}
\usepackage{amsmath}
\usepackage{amssymb}
\usepackage{booktabs} % For better table rules
\usepackage{multirow}
\usepackage{xcolor} % For text coloring
\usepackage{colortbl} % For table cell coloring (if needed)
\usepackage{caption} % For better caption control
\usepackage{subcaption} % For subfigures/subtables if needed
\usepackage{natbib} % For \[cite]p
\usepackage{hyperref} % For clickable references/links
\usepackage{float} % For [H] placement specifier
\usepackage{tabularx} % For flexible width tables
\usepackage{enumitem} % For customized lists
\usepackage{geometry} % Adjust margins if needed
\usepackage{tikz} % For potential diagrams if needed
\usepackage{pgfplots} % For plotting directly in LaTeX if needed
\usepackage{siunitx} % For consistent number/unit formatting
\usepackage{cleveref} % For smart cross-referencing (\cref)

\definecolor{forestgreen}{rgb}{0.13, 0.55, 0.13}

\usepackage{hyperref}

\usepackage[T1]{fontenc}
\usepackage[utf8]{inputenc}
\usepackage{microtype}
\usepackage{inconsolata}
\usepackage{graphicx}

\title{Quagmires in SFT-RL Post-Training: When High SFT Scores Mislead and What to Use Instead}

\author[1,2,*]{Feiyang Kang}
\author[1]{Michael Kuchnik}
\author[1]{Karthik Padthe}
\author[3]{Marin Vlastelica}
\author[2]{Ruoxi Jia}
\author[1]{Carole-Jean Wu}
\author[1]{Newsha Ardalani}

\affiliation[1]{FAIR at Meta}
\affiliation[2]{Virginia Tech}
\affiliation[3]{Meta}

\contribution[*]{Work done at Meta}
\abstract{In post-training for reasoning Large Language Models (LLMs), the current state of practice trains LLMs in two independent stages: Supervised Fine-Tuning (SFT) and Reinforcement Learning with Verifiable Rewards (RLVR, shortened as ``RL'' below). In this work, we challenge whether high SFT scores translate to improved performance after RL. We provide extensive counter-examples where this is not true. We find high SFT scores can be biased toward simpler or more homogeneous data and are not reliably predictive of subsequent RL gains or scaled-up post-training effectiveness. In some cases, RL training on models with improved SFT performance could lead to substantially worse outcome compared to RL on the base model without SFT. We study alternative metrics and identify generalization loss on held-out reasoning examples and Pass@large k performance to provide strong proxies for the RL outcome. We trained hundreds of models up to 12B-parameter with SFT and RLVR via GRPO and ran extensive evaluations on 7 math benchmarks with up to 256 repetitions, spending $>$1M GPU hours. Experiments include models from Llama3, Mistral-Nemo, Qwen3 and multiple state-of-the-art SFT/RL datasets. Compared to directly predicting from pre-RL performance, prediction based on generalization loss and Pass@large k achieves substantial higher precision, improving $R^2$ coefficient and Spearman's rank correlation coefficient by up to 0.5 (2x). This provides strong utility for broad use cases. For example, in most experiments, we find SFT training on unique examples for a one epoch underperforms training on half examples for two epochs, either after SFT or SFT-then-RL; With the same SFT budget, training only on short examples may lead to better SFT performance, though, it often leads to worse outcome after RL compared to training on examples with varying lengths. This work develops an enhanced evaluation tool that will be open-sourced.}

\date{\today}
\correspondence{Feiyang Kang at \email{fyk@vt.edu}, Newsha Ardalani \email{new@meta.com}}

\begin{document}

\maketitle

\begin{figure}[h!]
    \centering
        \includegraphics[width=\textwidth]{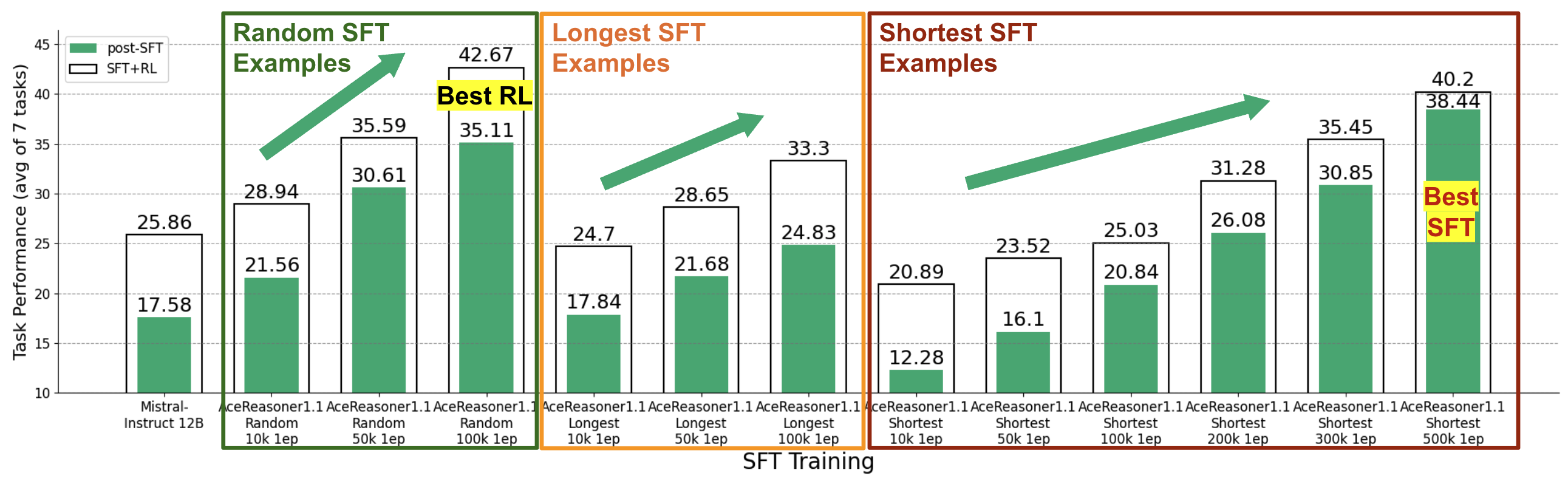}
        \vspace{-0.5em}
    \caption{\small{Mistral-NeMo-12B-Instruct undergone SFT-RL with SFT examples from AceReasoner1.1-SFT dataset and RLVR via GRPO on DeepScaleR dataset. Reporting Pass@1 performance averaged over 7 math benchmarks. When training on \textcolor{Green}{Random}/\textcolor{Orange}{Longest}/\textcolor{Maroon}{Shortest} SFT examples, \textit{the final performance after RL increases at different rates than the SFT performance.} Model with the best SFT performance is not the one with the best final performance after RL. Post-SFT and SFT+RL performance correlate, though, optimizing post-SFT performance might not optimize the final performance after RL.
}\normalsize }\vspace{-1em}\label{fig:teaser-intro}
\end{figure}

The evolution of Large Language Models (LLMs) has seen a significant focus on enhancing their reasoning abilities, a process heavily reliant on post-training \citep{wen2025light}. This phase refines pre-trained models, adapting them for complex, multi-step tasks like mathematics, logic, and code generation, leading to the emergence of Large Reasoning Models (LRMs) \citep{kumar2025llm}. The open-sourced DeepSeek R1 achieved phenomenal success in pushing forward the frontier of LLM's reasoning capabilities \citep{guo2025deepseek}. Its new post-training paradigm, Reinforcement Learning with Verifiable Rewards (RLVR) via Group Relative Policy Optimization (GRPO) \citep{liu2024deepseek,shao2024deepseekmath}, has shown substantial improvements on top of previous post-training methods.
Following DeepSeek R1's practice, current works typically conduct SFT before RL, assuming models with better performance after SFT will ultimately be better after RL \citep{liu2025acereason,wen2025light}. In industrial practice, these post-training stages are often distributed among different teams, with SFT and RL handled by separate groups, each optimizing for their own performance metrics \citep{chen2025acereason,meta2025llama}. This process relies on the intuition that a model with stronger SFT performance will yield better outcomes after RLVR \citep{liu2025acereason}. With efforts and resources being poured in improving post-training paradigms and data recipes, also escalating are the debates on \textit{whether SFT helps or hurts the subsequent RL training}.

In this setup, post-training strategies and data are often designed either for SFT or RL, but not jointly. In practice, SFT and RL are often conducted sequentially (e.g., \citet{rastogi2025magistral}). SFT data is usually selected to maximize evaluation performance after SFT \citep{zhang2025best,ye2025limo}, and the best-performing SFT models are believed to also yield stronger performance after subsequent RL. However, this assumption is often flawed. Over-training during SFT, for instance, can constrain the model's behavior and limit the exploration crucial for effective RL \citep{chen2025sft,wang2025beyond}. For example, we find training on repeated examples for up to 8 epochs leads to better SFT performance than training on the same data for 2 epochs (4x compute) but yields visibly worse outcome after RL (Figure \ref{fig:teaser-epoch}, left). On the contrary, \citet{cen2025behavior} shows SFT training on manually crafted ``exploratory'' examples, despite leading to a lower performance after SFT, helps achieve better final outcome after RL. This leads to a critical gap in the current practice:

% However, this assumption is often flawed. Over-training during SFT, for instance, can constrain the model's behavior and limit the exploration crucial for effective RL [cite]. For example, we find training on repeated examples for up to 10 epochs leads to better SFT performance than training on the same data for 3 epochs (3.3x compute) but yields visibly worse outcome after RL (Figure 6). This leads to a critical gap in the current practice:

\textit{An SFT-trained model with the best evaluation performance may not be the best candidate for subsequent training with RLVR (e.g., Figure \ref{fig:teaser-intro}).}

\begin{figure}[h!] 
    \centering
    
    % % \hfill
    % \hspace{-1em}
    \begin{subfigure}[b]{0.55\textwidth}
        \vspace{-0.5em}\includegraphics[width=\textwidth]{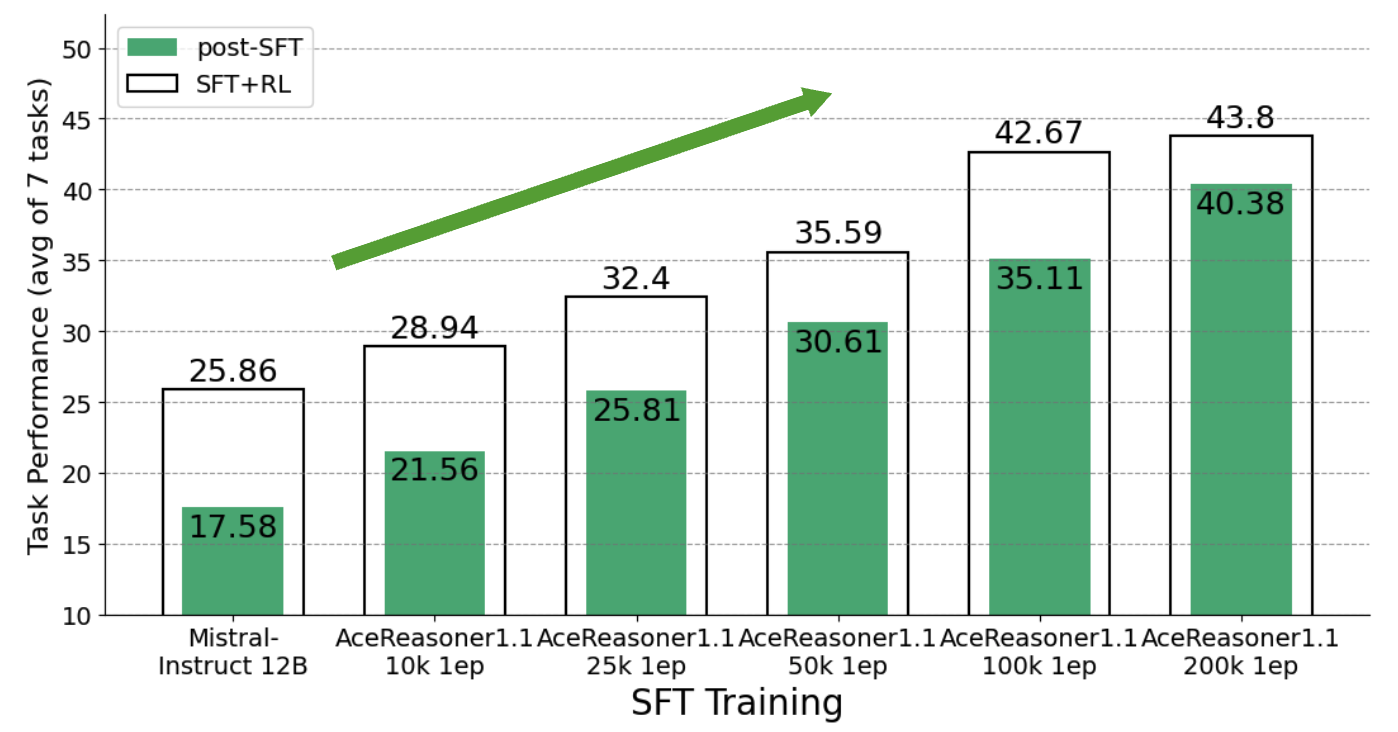}% \vspace{-0.5em}
        \caption{Mistral-NeMo-12B-Instruct}
    \end{subfigure} 
        \begin{subfigure}[b]{0.42\textwidth}
        \includegraphics[width=\textwidth]{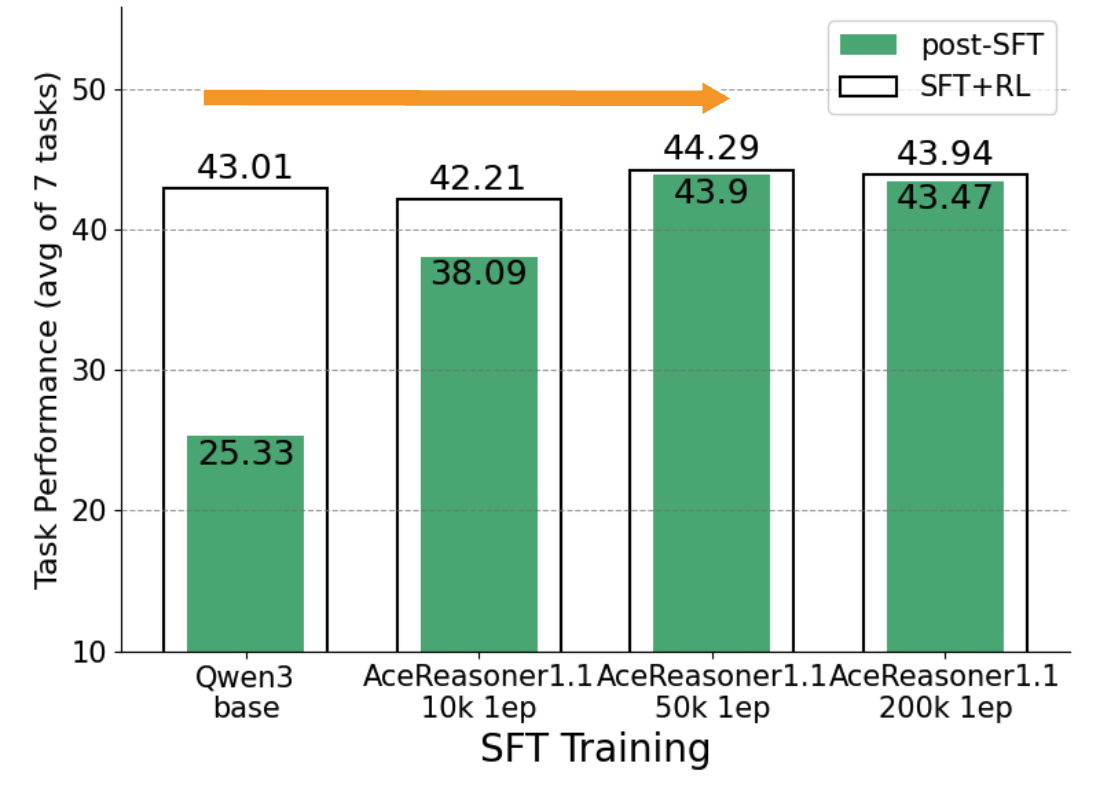}\vspace{-0.2em}
        \caption{Qwen3-4B-base}
    \end{subfigure}
     \caption{\small{Both models undergone SFT-RL with SFT examples from AceReasoner1.1-SFT dataset and RLVR via GRPO on DeepScaleR dataset. Reporting Pass@1 performance averaged over 7 math benchmarks. \textit{Even with identical post-training procedures, different models may respond vastly different.} With increasing SFT examples, Mistral's (left) post-SFT performance and final performance after RL both increase. Yet, for Qwen3 models (right), the post-SFT performances appear uncorrelated with the final performance after RL, where the latter remains the same despite the substantially improved SFT performance.} \normalsize}
    \label{fig:teaser-vol}
\end{figure}

When the final RLVR performance is unsatisfactory, it becomes challenging to attribute the failure to either the RL stage or a non-ideal SFT starting point. This misalignment can cause friction and overhead between teams. Furthermore, the high computational cost of RL training and long pipelines, especially in agentic use cases, makes end-to-end tuning across the SFT-RL stages prohibitively expensive \citep{toledo2025ai}. Early stopping during RL is also generally ineffective, as the model with the fastest initial improvement may not achieve the highest final performance \citep{liu2025acereason}. Even with identical post-training procedures, different models may respond vastly different (Figure \ref{fig:teaser-vol}). Consequently, a significant gap remains in our ability to reliably predict RLVR outcomes.

This work centers on addressing this predictability problem. We ask the following research questions:
\begin{itemize}
    \item[RQ1:] \textit{Do models with better pre-RL performance always lead to better outcomes after RLVR? If not, what are the failure modes? (Section \ref{sec:trap})}
    \item[RQ2:] \textit{What are effective SFT paradigms and data recipes when considering subsequent RLVR training? Can we determine the suitability of an SFT model before committing to the expensive RL stage? (Section \ref{sec:metric})}
\end{itemize}

To tackle these questions, we first examine the relationship between pre-RL performance and post-RL outcomes across various SFT training paradigms and data recipes. While we often observe some extents of correlation between post-SFT performance and final outcome after RL, we identify cases where the trends visibly diverge. For instance, training on the same dataset for more epochs may significantly boost post-SFT performance but diminishes the potential for improvement during RL, sometimes resulting in lower final performance (e.g., Figure \ref{fig:teaser-epoch}). Similarly, training on simpler, shorter reasoning examples improves pre-RL performance quickly, but these models gain much less from the subsequent RL stage. These findings are particularly concerning given that many SFT data selection methods favor simpler or more homogeneous examples \citep{zhang2025best,yu2025rip}.

Next, we identify more reliable predictors for RL success. We find that as SFT proceeds, an eventual increase in validation loss is strongly correlated with performance improvements in the later RL stage. Furthermore, since the RL objective is to compress Pass@k performance into Pass@1 (i.e., maximize expected reward), we investigate using Pass@k at a large k as a predictor. We conduct extensive empirical validation using Llama3-8B \citep{grattafiori2024llama}, Mistral-Nemo-12B \citep{mistral2024nemo}, and Qwen3-4B-base \citep{yang2025qwen3} models on state-of-the-art SFT datasets like Llama-Nemotron \citep{singhal2025llama} and AceReasoner1.1 \citep{liu2025acereason} and different RL datasets. Our results demonstrate that these new metrics can reliably predict the outcome of RLVR, improving $R^2$ coefficient and Spearman's rank correlation coefficient by up to 0.5 (2x), providing strong utility for broad use cases. 
For example, in most experiments, we find SFT training on unique examples for a one epoch underperforms training on half examples for two epochs, either after SFT or SFT-then-RL; With the same SFT budget, training only on short examples may lead to better SFT performance, though, it often leads to worse outcome after RL compared to training on examples with varying lengths. These can be captured by the proposed predictors, but not from the post-SFT performance.

To address the limitations in available tools, we developed an enhanced tool for more convenient and reliable evaluation of reasoning models, which will be open-sourced in contribution to the community.

\section{Related Works}

% Since OpenAI's o1 started the paradigm of scaling up reasoning during inference time, contemporary LLMs enters the era of solving problems via reasoning. 

% Post-training for reasoning LLMs typically consists of two or three stages: a) Supervised Fine-Tuning (SFT), b) an optional Direct Preference Optimization (DPO) stage, and c) Reinforcement Learning with Verifiable Rewards (RLVR). 

The research landscape for reasoning post-training and data strategies is fast evolving and in its early days. In a typical setup, post-training for reasoning LMs conducts SFT and RL sequentially, which has been reported to work better than only conducting SFT or RL \citep{rastogi2025magistral}. 
Viewpoints in many impactful works can be inconsistent or even contradicting: 
“Initial ‘cold-start’ SFT is necessary for subsequent RL” (DeepSeek-R1 technical report, \citet{guo2025deepseek});
“over-SFT may constrain subsequent RL” (Llama-4 technical report, \citet{meta2025llama}); 
“SFT generalizes poorly and RL without SFT does better” \citep{chen2025sft}, showing prominent gaps in characterizing post-training dynamics and the role of each stage. The lack of predictability in the post-training outcome poses a major blocker for optimizing training paradigms or data recipes.

% We hypothesize that this is due to their particular configurations for post-training or/and evaluation.

\subsection{Post-training for Reasoning: SFT-then-RLVR paradigm}
% Reasoning language models (RLMs) became extremely popular following the success of DeepSeek R1. Performance on reasoning tasks (math/code) improved substantially via large-scale reinforcement learning (RL) on problems with verifiable answers (aka reinforcement learning with verifiable rewards, RLVR). RL discovers solution traces on its own, not relying on expert-curated examples, and it turns out to be more effective.

Post-training for reasoning LLMs typically consists of two or three stages: a) \textbf{Supervised Fine-Tuning (SFT)}, b) an optional \textbf{Direct Preference Optimization (DPO)} stage, and c) \textbf{Reinforcement Learning with Verifiable Rewards (RLVR)} \citep{lambert2024tulu}. SFT serves as the ``cold-start'' phase, providing the model with a strong initial policy by exposing it to high-quality reasoning chains \citep{guo2025deepseek}. The model is trained on problems with high-quality solutions sourced from the frontier models. DPO fixes/strengthens targeted behaviors (e.g., precise instruction following in math/logic derivations) assessed important for effective reasoning, which is more subjective and often optional \citep{lambert2024tulu}. RL further improves the model’s reasoning and problem-solving capability. This allows the model to explore the solution space more broadly than SFT alone, discovering novel and more robust reasoning paths. 

While the sequential SFT-then-RL pipeline is dominant, researchers have explored alternative paradigms to more tightly integrate or unify these learning stages. Efforts include iterate or interleave SFT and RL \citep{meta2025llama}, gradually shift from SFT to RL while increasing task difficulty \citep{yang2025qwen3}, or directly unify the objectives of SFT and RL \citep{xu2025kdrl}. Though these are promising research directions, they come with their own complexities and have not yet universally replaced the SFT-then-RL paradigm, which remains a robust and widely-adopted industry standard. Many important issues regarding the stability, data requirements, and effectiveness of these unified methods remain to be solved. Our work, therefore, focuses on improving the predictability and efficiency of the prevailing SFT-then-RL pipeline, providing practical tools that are immediately applicable to current state-of-the-art workflows.

\subsection{Recent Advancements and Current Challenges}
In post-training for reasoning, SFT data is usually selected to maximize evaluation performance after SFT \citep{li2025naturalthoughts,ye2025limo}, and the best-performing SFT models are believed to also yield stronger performance after subsequent RL. Significant research effort is now focused on more sophisticated selection and curation strategies for SFT data. Techniques range from filtering for complexity and diversity to generating synthetic data that covers a wider range of reasoning structures \citep{rastogi2025magistral,yuan2025naturalreasoning,ye2025limo,abdin2024phi}. Some methods propose selecting data points based on their difficulty and influence, aiming to find a subset of examples that provides the strongest learning signal \citep{muennighoff2025s1}. Current efforts prioritize scaling up SFT training on existing models, leading to new SOTA performance on reasoning tasks for those models \citep{guha2025openthoughts}. A significant challenge is that standard SFT performance metrics, such as average accuracy on benchmarks, are not always predictive of post-RL success. This creates a critical gap between the optimization target of the SFT stage and the final performance of the model.

Several issues contribute to this gap. First, models can overfit to the specific patterns and artifacts present in the SFT dataset, leading to poor generalization during the exploration phase of RL \citep{chen2025sft}. Furthermore, naively collecting or generating data can lead to datasets that lack diversity in reasoning strategies or are skewed toward simpler problems, causing the SFT-trained model to develop biases that stifle exploration in the subsequent RL stage \citep{guha2025openthoughts}. The landscape is further fogged by the recently reported data 
contamination issues \citep{wu2025reasoning}. The results from these models have served as the basis for many research findings.

% If the SFT data is homogeneous or biased towards simpler problems, the model may learn a narrow policy that performs well on similar examples but fails to provide a good foundation for RL to build upon, stifling exploration and leading to suboptimal final performance. 

The (lack of) predictability for final performance after RL from pre-RL models leads to quagmires for post-training. SFT teams may provide suboptimal RL learners. It creates frictions between post-training teams owning different SFT and RL stages and chaos in optimizing the training paradigm/data recipes, adding overheads on the model development and hindering productivity. It calls for new tools that better characterize the post-training dynamics and predictive of the RL outcome. This will have profound impact on broad downstream fields--research and applications alike--from improving SFT data curation, search for the next post-training paradigm, to RL for non-verifiable tasks, etc.

\section{The SFT Metric Trap}\label{sec:trap} Previous works, from SFT data selection to RL training methodologies, have often operated under a common assumption. They \textit{implicitly} assume or \textit{explicitly} argue that models exhibiting better post-SFT performance will consistently yield superior final outcomes after subsequent reinforcement learning \citep{rastogi2025magistral,liu2025acereason}. This assumption has justified the widespread practice of optimizing the SFT and RL stages in isolation, with teams or processes focusing on maximizing SFT evaluation metrics as a primary goal. However, the separation of SFT and RL optimization can lead to a widening gap in reasoning post-training, where improvements in the initial stage do not translate to the final stage. This motivates us to ask two fundamental questions: 

\begin{itemize}
\item \textit{Do models with better pre-RL performance always lead to better outcomes after RLVR?}
\item \textit{If not, what are the failure modes?}
\end{itemize} 

To investigate these questions, we design experiments across two representative scenarios that reflect common practices and research directions in the field: a ``dataset-level'' analysis and an ``instance-level'' analysis. In \textbf{Dataset-Level Scenarios}, SFT examples are drawn from the same data distribution, but we vary the amount of unique samples and the training paradigm (e.g., learning rate, number of epochs); In \textbf{Instance-Level Scenarios}, we consider training on different datasets while keeping the training pipeline fixed (i.e., using the same model and training paradigm). This setup is primarily concerned with SFT data selection and curation, examining whether strong SFT performance on a given dataset transfers to the final outcome after RL.

\subsection{Dataset-Level Scenarios} \label{sec:dataset_level} 
In this scenario, we draw SFT examples from the same underlying data distribution but vary the training configuration, such as the number of unique samples/training epochs/learning rate. This setup is highly relevant to industrial practices where SFT and RL are often handled by different teams. In current practices, the number of training epochs is a design choice often determined by practical factors such as data availability or compute budget. Specifically, when the amount of training samples is a more prominent constraint (such as domains with limited high-quality examples), repeating for more epochs on the data may be preferred to improve post-SFT performance. On the contrary, if data is abundant relative to the allocated compute budget (for this domain/capability), current practices (such as \citet{singhal2025llama}) may prefer to train for just a single epoch on unique examples.

\begin{wrapfigure}{R}{0.48\textwidth}\vspace{-0em}
\begin{minipage}{0.49\textwidth}
\vspace{-1em}
    \scalebox{0.27}{\includegraphics{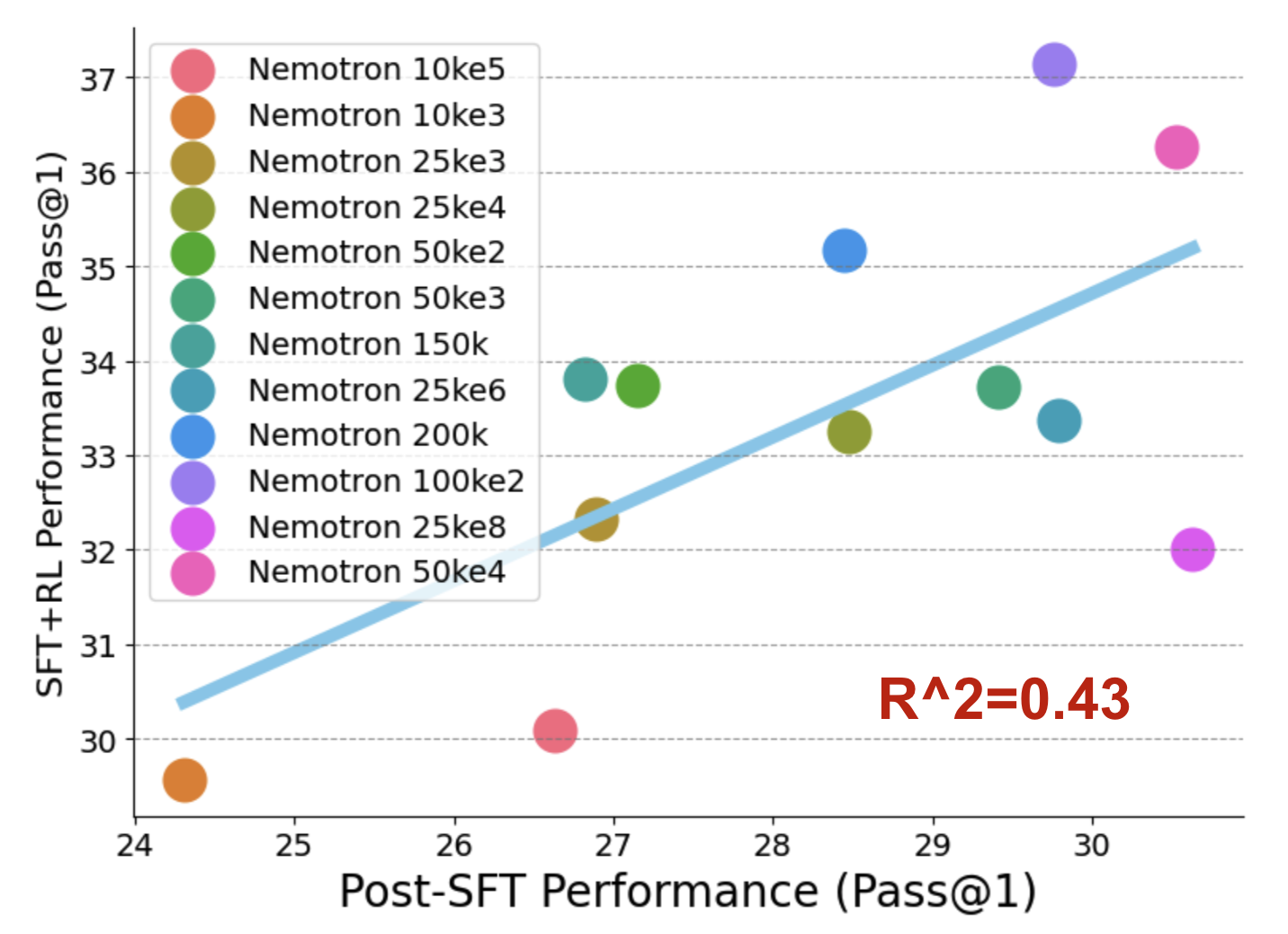}}\vspace{-0.0em}
     \caption{\small{
     Llama3-8B-Instruct undergone SFT-RL with SFT examples from Llama-Nemotron-SFT dataset and RLVR via GRPO on MATH dataset (train-split). Reporting Pass@1 performance averaged over 7 math benchmarks. Linear fit between post-SFT performance and final outcome after RL. The two performance correlates with $R^2=0.43$, indicating \textit{post-SFT performance explains only 43\% of variation in the final outcome after RL and the remaining gaps are prominent.}}\normalsize}
    \label{fig:SFT-epoch-baseline}\vspace{-1.5em}\hspace{-5em}
\end{minipage}
\end{wrapfigure}

\begin{figure}[h!] 
    \centering
    % % \hfill
    % \hspace{-1em}
    \begin{subfigure}[b]{0.55\textwidth}
        \vspace{-0.5em}\includegraphics[width=\textwidth]{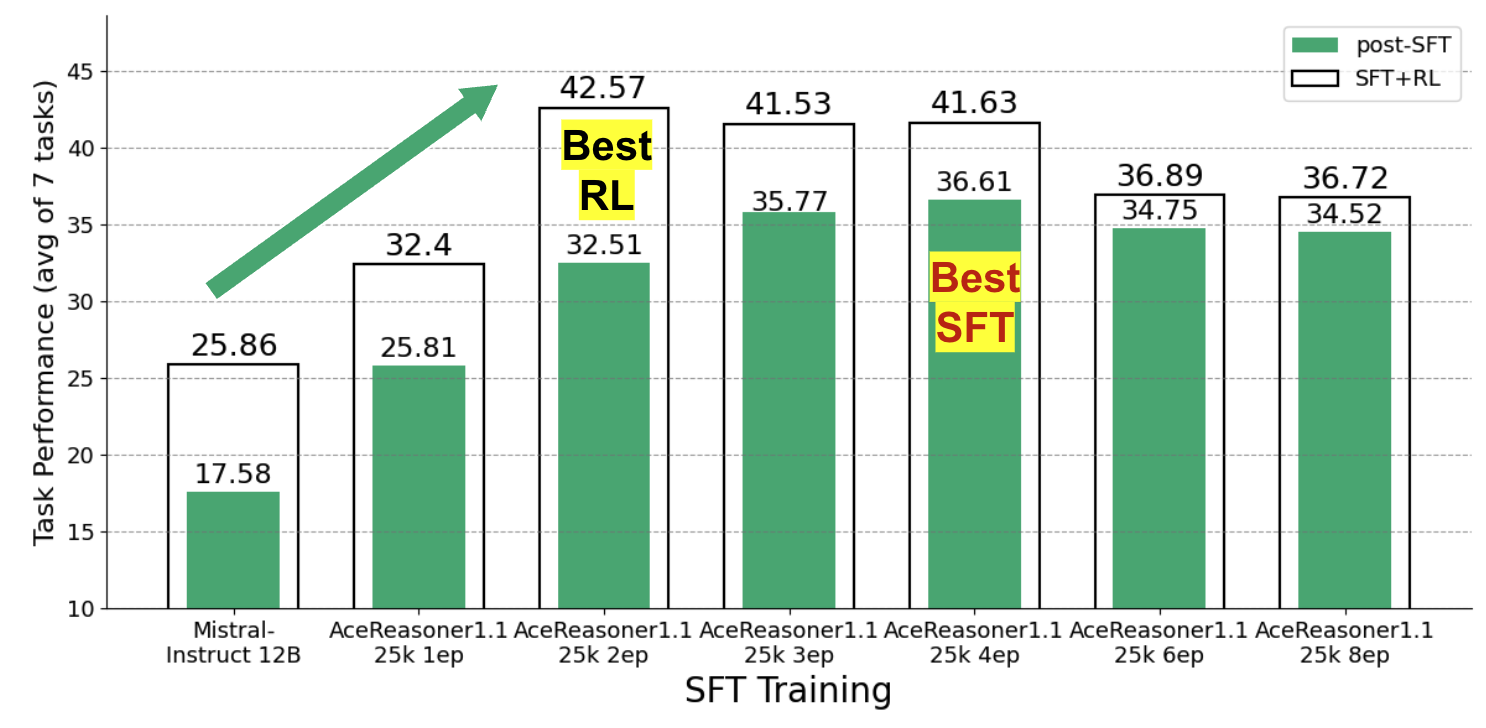}% \vspace{-0.5em}
        \caption{Mistral-NeMo-12B-Instruct}
    \end{subfigure}
    \begin{subfigure}[b]{0.43\textwidth}
        \includegraphics[width=\textwidth]{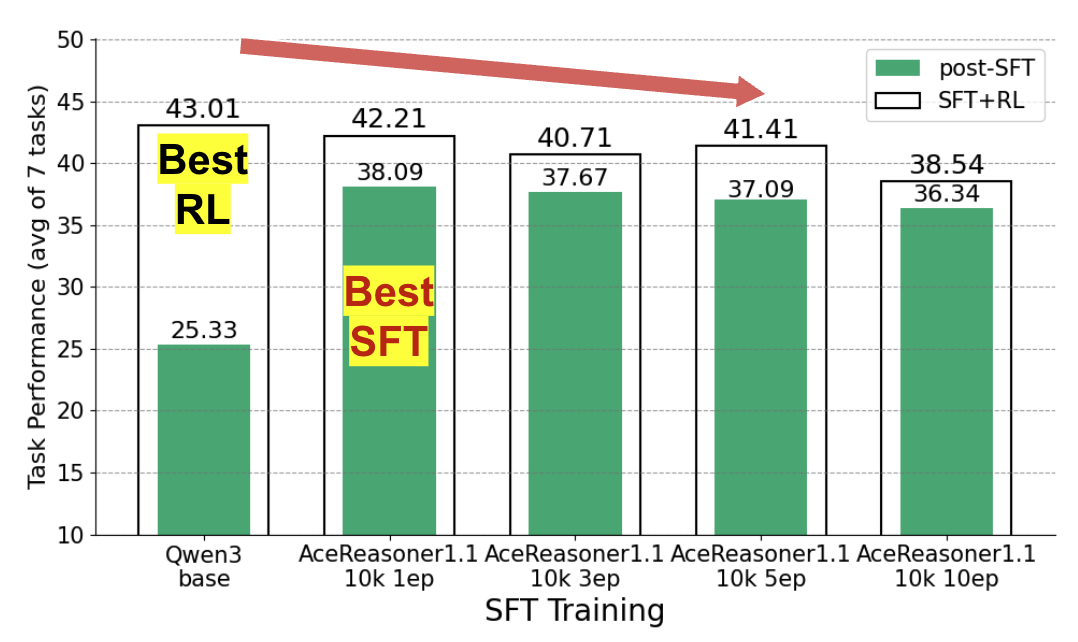}\vspace{-0.2em}
        \caption{Qwen3-4B-base}
    \end{subfigure}
      \caption{\small{Both models undergone SFT-RL with SFT examples from AceReasoner1.1-SFT dataset and RLVR via GRPO on DeepScaleR dataset. Reporting Pass@1 performance averaged over 7 math benchmarks. When repeating SFT for more epochs on the same data, Mistral's (left) SFT continues to improve with up to 4 epochs where the final performance after RL saturates after 2 epochs. Qwen3's (right) final performance after RL degrades with SFT training, though, these models' post-SFT performance is substantially higher than the base model. \textit{Both cases show clear divergence between post-SFT performance and final performance after RL. Here, optimizing post-SFT performance will be suboptimal or ineffective for improving the final model.}\normalsize}} 
    \label{fig:teaser-epoch}
\end{figure}

In these cases, the training paradigm is determined \textbf{heuristically} where the only optimizable target is the post-SFT performance. Surprisingly, we identified both practices to be \textbf{suboptimal}. We found that post-SFT performance often improves stably when training for more epochs--even with excessive overtraining. But models overtrained during SFT show decreasing potentials for the subsequent RL. Typically, the model with the best final performance after RL is not the one with the best post-SFT performance.  Further, with the same compute budget for SFT, training on more data for one epoch typically leads to visibly lower post-SFT performance compared to training on less data for a few more epochs, and the final performance after RL remains underperforming. A concrete example is provided in Figure \ref{fig:teaser-epoch}. High SFT scores can be biased toward \textit{homogeneous or repeated examples} and are not reliably predictive of subsequent RL gains. 

This mismatch between post-SFT and post-RL performance is not directly visible from post-SFT models. As shown in Figure \ref{fig:SFT-epoch-baseline} where we fit a linear function between post-SFT and post-RL performance, these two performance correlates with $R^2=0.43$, indicating post-SFT performance only explains 43\% of variation in the final outcome after RL whereas the gaps remain evident.

\subsection{Instance-Level Scenarios} \label{sec:instance_level} % ... content for instance-level experiments ...

In this scenario, we fix the model and the training configurations but vary the SFT datasets. This setup is primarily concerned with SFT data selection and curation, examining whether the strong SFT performance promised by a particular dataset transfers to strong final performance after RL. For instance, state-of-the-art data selection methods are often prone to selecting examples that are more ``natural'' or easier for the model to learn \citep{zhang2025best,yu2025rip}. While this simpler data may allow the model to achieve high SFT metrics more quickly, we question whether this comes at the cost of learning more difficult or advanced reasoning capabilities that are crucial for downstream success.
We identified similar gaps between post-SFT performance and final outcome after RL. Visualizing representative examples in Figure \ref{fig:teaser-intro}, high SFT scores can be biased toward \textit{simpler examples} and are not reliably predictive of subsequent RL gains or scaled-up post-training effectiveness. For example, \textit{training on shortest examples led to faster performance improvements} than training on randomly sampled examples during SFT. These shorter examples are closer to the model’s original generations and easier to learn, though, these are not best examples for the model to gain reasoning capabilities in preparation for RL. The final performance after RL is significantly lower. These gaps are not directly captured in the post-SFT performance.

\section{Proposed Metrics towards More Reliable Predictions}\label{sec:metric}

\subsection{Generalization Loss on Validation Examples}

% surprisingly, we found performance for SFT (average Pass@1) to improve stably even with excessive over-training, e.g., training for many epochs or/and with much larger learning rates (Figure 1, i/ii)

% even when
% generalization loss on validation examples elevates and eventually flare up, indicating strong over-fitting

% this over-fitting does not hurt SFT Pass@1

% we hypothesize the generalzation loss is composed of 
% heterogenious components

% to patterns
% verbal patterns

%  reasoning capabilities
% improving 

% the loss of verbal diversity does not affect SFT performance, though, it 

% observation:

% clear divergence

% under-training, over-training

% % entropy

\begin{figure}[h!]
    \centering\vspace{-0em}
        \includegraphics[width=0.9\textwidth]{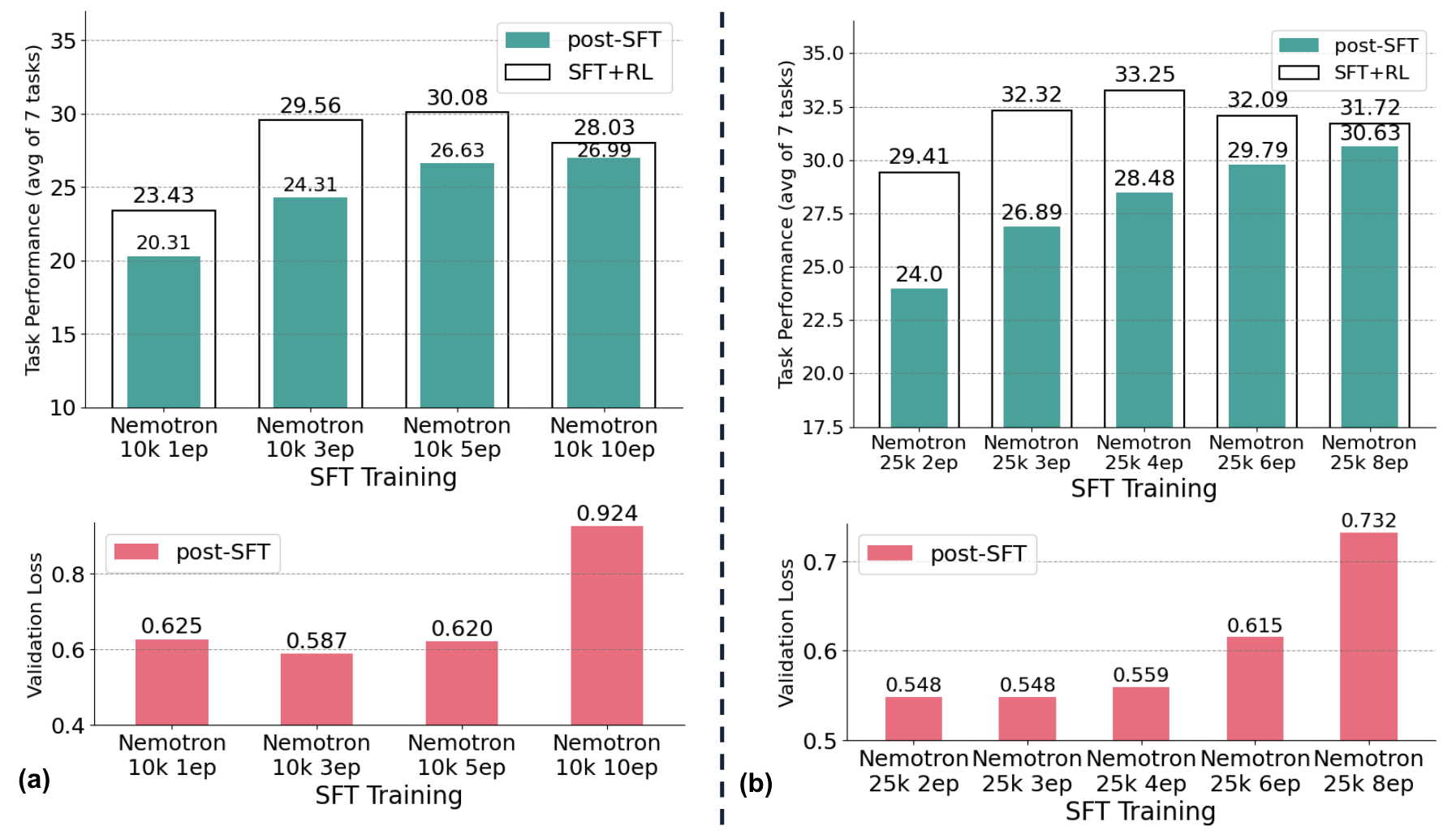}
        \vspace{-0.0em}
    \caption{\small{Llama3-8B-Instruct undergone SFT-RL with SFT examples from Llama-Nemotron-SFT dataset and RLVR via GRPO on MATH dataset (train-split). Reporting Pass@1 performance averaged over 7 math benchmarks and generalization loss on the validation set of SFT data. We identify generalization loss after SFT to be a viable indicator for the model's RL potential. While repeating training for more epochs, together with the improving post-SFT performance, we observe the generalization loss on validation examples to elevate and eventually flare up, indicating strong over-fitting. \textit{This generalization loss shows strong correlation with the further performance gain during the subsequent RL, allowing prediction for the final outcome after RL.}}\normalsize}\label{fig:gen-loss-epoch}
\vspace{-0em}
\end{figure}

During the investigation above, we identified a counterintuitive pattern in which post-SFT performance improves stably when training for more epochs whereas the overtrained models show decreased potentials during the subsequent RL. The best final performance after RL is not usually achieved on models with the best post-SFT performance. 
To be able to optimize the final outcome on the given training examples, one needs to optimize the SFT training paradigm based on the predicted final outcome after RL. We materialize this insight and identify generalization loss after SFT to be a viable indicator of the model's potential during the subsequent RL. While repeating training for more epochs, together with the improving post-SFT performance, we observe the generalization loss on validation examples to elevate and eventually flare up, indicating strong over-fitting. This generalization loss shows strong correlation with further performance gains during subsequent RL, allowing prediction for the final outcome after RL (Figure \ref{fig:gen-loss-epoch}). When using it in practice, after conducting SFT training with different numbers of examples and epochs, we can immediately rule out post-SFT models with both lower performance and higher generalization loss as they will likely remain underperforming after the subsequent RL, facilitating determination of the best SFT training paradigm.

\subsection{Pass@k Accuracy Evaluated at Large k}
The objective of RLVR via GRPO is to maximize expected reward, which explicitly optimizes the Pass@1 accuracy on the RL tasks. GRPO only progresses when at least one of the responses for the RL task is correct. Recent works argue that GRPO compresses Pass@k accuracy into Pass@1 \citep{yue2025does}, and empirical evidence appears to support the argument showing GRPO mostly improves average Pass@1 accuracy on tasks where the original model achieves an above-zero accuracy \citep{liu2025acereason}. Though it remains debatable whether GRPO discovers new solution traces beyond the capabilities of the original model \citep{liu2025prorl}, all these analyses and findings suggest RLVR dynamics during GRPO training to be strongly coupled with the original models Pass@k accuracy. \citet{hu2023predicting} pioneers in using the Pass@high metric to study the scaling of task performance. The authors argue that Pass@k provides finer resolution to the Pass@1 metric and better captures the underlying dynamics. Acting on this intuition, we consider Pass@k performance of the post-SFT model, especially with large k, as a candidate metric for predicting its final outcome after the subsequent RL. When using it in practice, after SFT training, we evaluate Pass@k performance on the post-SFT models with different values of k. For efficient implementation, we leverage the following formula which provides unbiased estimations for Pass@k accuracies for all $k\leq n$ \citep{brown2024large}, 
\begin{equation}
    \text{Pass}@k = \mathbb{E}\left(1-\frac{\binom{n-c}{k}}{\binom{n}{k}}\right)
\end{equation}
where integer $n$ denotes the total number of responses generated for the task, integer $k$ denotes the target value for k Pass@k, and integer $c$ denotes the number of correct responses for the task, respectively. 

We consider the Pass@large k performance as the indicator for the final outcome after RL and deem the post-SFT model with the best Pass@large k performance to have the best Pass@1 performance after RL. The best post-SFT model can be determined without needing to conduct any actual RL run. % A design choice is the value of k. A smaller k may not be able to capture sufficient signals whereas the model's performance could saturate at a large k, both leading to elevated noise in estimation. Besides, evaluating Pass@k accuracy requires repeating the evaluation for at least k times, which becomes notably expensive at large values of k. A recent work [llm monkeys] finds Pass@k accuracy on a benchmark consisting multile questions scales in a log-linear function with the value of k, and provides theotical groundings for this formula in a range of scenarios. This enables estimating Pass@k accuracy at a larger k from that of estimated smaller k by extropolating the scaling function.

\section{Empirical evaluations}
\subsection{Setup}
We conduct three sets of experiments with SFT-RL post-training. On \textbf{Llama3-8B-Instruct} models, we conduct SFT training with examples from Llama-Nemotron dataset (where we only select math samples with responses generated by QwQ-32B \citep{qwq32b} or DeepSeek R1 \citep{guo2025deepseek}, hereinafter the same) and RL training on MATH dataset (train-split) \citep{hendrycks2021measuring}; on \textbf{Mistral-Nemo-12B-Instruct} and Qwen3-4B-base,  we conduct SFT training with examples from AceReasoner1.1-SFT dataset and RL training on DeepScaleR dataset \citep{deepscaler2025}. For all models, we conduct RL training for 3 epochs where each run  takes up to 5 days. We repeat RL training for 4+ runs on each data recipe and training paradigm, conduct 4+ evaluations on different checkpoints across RL training run, and report the best performance for the model. We evaluate task performance on 7 math benchmarks, \textbf{MATH-500} \citep{hendrycks2021measuring}, \textbf{AIME 1983-2024} \citep{aime_1983_2024}, \textbf{GSM8k} \citep{cobbe2021gsm8k}, \textbf{AIME 2025} \citep{aime_2025}, \textbf{AMC} \citep{amc}, \textbf{Olympiad} \citep{he2024olympiadbench}, \textbf{Minerva} \citep{lewkowycz2022solving}, and report model performance as Pass@1 accuracy averaged over 64 repetitions and across 7 tasks. For the proposed predictors, we evaluate the generalization loss on the validation set of the SFT data and Pass@64 accuracy averaged over 256 repetitions. Experiments spent $>$1M GPU hours on NVIDIA A100. Please refer to Appendix \ref{app:setup} for additional details. Shown in Figure \ref{fig:teaser-vol} (right), in this setup, Qwen3-series models do not appear to benefit from state-of-the-art SFT datasets, and models undergone different SFT training achieve considerably close final performance after RL. Since this work focuses on studying the impact of different SFT training on the subsequent RL, we present these results as qualitative examples instead (deferred to Appendix \ref{app:vis}).

Following the categorization above, we organize experiments in two major scenarios: \textbf{dataset-level} prediction, and \textbf{instance-level} prediction. In dataset-level prediction experiments: we conduct SFT training for the base model on samples from math reasoning datasets with different training paradigms (varying number of examples and epochs). In instance-level prediction experiments: we first create diverse different curated SFT datasets by selecting the shortest/longest subsets, random samples, or their different mixtures (samples are shown in Appendix \ref{app:sample}). Then, we conduct SFT training for the base model on samples from each curated dataset with the same training paradigms (one epoch). We consider two primary metrics measure prediction performance, \textbf{Coefficient of determination ($R^2$)} \citep{pearson1909determination}, and \textbf{Spearman's rank correlation coefficient (Spearman)} \citep{zar1972significance}. Specifically, $R^2$ measures the proportion of variation in the prediction variable (final performance) that is unexplained by the predictor, examining the accuracy of prediction on the final performance after RL. Spearman yields a number ranging from -1 to 1 that indicates how strongly two sets of ranks are correlated, which we use to examine the effectiveness in identifying post-SFT models that lead to the best final performance. Additional results can be found in Appendix \ref{app:res}.

\subsection{Use Case 1: Dataset-Level Prediction}
This use case focuses on optimizing the SFT training paradigm, a common dataset-level challenge. Given a fixed compute budget, practitioners must decide on the optimal trade-off between the volume of unique data and the number of training epochs, navigating the risks of under- and over-training. We test the predictive power of our proposed metrics against the baseline of using post-SFT Pass@1 accuracy. To examine the accuracy of prediction with $R^2$, we randomly select 50\% SFT models and fit a linear function between their post-SFT performance and final performance after RL. The fitted function is then used to predict the final performance of the other 50\% SFT models. We compare the predictions to their actual post-RL outcomes to compute $R^2$. We repeat the random sampling for 100 times and report the standard error.

\begin{table}[h!]
\centering
\caption{\small{Spearman's rank correlation between performance predicted from post-SFT models and the actual performance after RL. Both generalization loss and Pass@64 achieve notable margins over prediction from Pass@1, whereas averging the two prediction may or may not lead to better results.}\normalsize}\vspace{-0.0em}
\resizebox{1.0\linewidth}{!}{
\begin{tabular}{l|c|cc|c}
\toprule
\textbf{Spearman's Rank} & Prediction based on & Prediction based on & Prediction based on & Avg. Prediction from \\
\textbf{Correlation / Models} & SFT Pass@1 (avg. &  SFT Generalization&  SFT Pass@Large k & SFT Gen. Loss + \\
\textbf{ } & of 64) \textcolor{Orange}{\textbf{baseline}}&  Loss &   (k=64) &  Pass@Large k (64)\\
\midrule
Llama3-8B-Instruct                        & 0.75                  & 0.94 & 0.95 & \textbf{0.97} (\textcolor{Green}{+0.22})                 \\ 
Mistral-NeMo-12B-Instruct                        & 0.78  & 0.90 &  \textbf{0.92} (\textcolor{Green}{+0.14})                  & 0.90                \\ 
% % Mistral-12B-Instruct                        & 0.71  & 0.82 &  \textbf{0.88} (\textcolor{Green}{+0.17})                  & 0.85                \\ 
% \midrule
% Large Budget ($5\sim20$B tokens)     & 46.56 (-1.48)          & 27.86 (-0.25)          \\  
\bottomrule
\end{tabular}
}
\end{table}\vspace{-0.0em}
\begin{table}[h!]
\centering
\caption{\small{Measuring prediction accuracy with coefficient of determination ($R^2$). We randomly select 50\% SFT models and fit a linear function between their post-SFT performance and performance after RL, and use it to predict for the other 50\% SFT models. We repeat random sampling for 100 times and report standard errors.}\normalsize}\vspace{-0.0em}
\resizebox{1.0\linewidth}{!}{
\begin{tabular}{l|c|cc|c}
\toprule
\textbf{Coefficient of} & Prediction based on & Prediction based on & Prediction based on & Avg. Prediction from \\
\textbf{determination ($R^2$)} & SFT Pass@1 (avg. &  SFT Generalization&  SFT Pass@Large k & SFT Gen. Loss + \\
\textbf{/ Models} & of 64) \textcolor{Orange}{\textbf{baseline}}&  Loss &   (k=64) &  Pass@Large k (64)\\
\midrule

Llama3-8B-Instruct                        & 0.57 \tiny$\pm$ 0.29       & 0.88 \tiny$\pm$ 0.09 & 0.87 \tiny$\pm$ 0.10          & \textbf{0.94} \tiny$\pm$ 0.04 \normalsize(\textcolor{Green}{+0.37})                \\ 

Mistral-NeMo-12B-Instruct                        & 0.29 \tiny$\pm$ 0.38       & \textbf{0.79} \tiny$\pm$ 0.26 \normalsize(\textcolor{Green}{+0.50})  & 0.57 \tiny$\pm$ 0.32          & 0.72 \tiny$\pm$ 0.24              \\

\bottomrule
\end{tabular}
}
\end{table}

\begin{tcolorbox}[title=\text{Takeaway 1: Dataset-level Prediction},colback=lightgray!30!white,colframe=black]
\begin{itemize}[leftmargin=*]
    \item Both generalization loss and Pass@large k are effective predictors for post-RL performance when optimizing SFT training configurations on a single dataset, providing higher-accuracy estimates that help guide decisions and save significant compute.
\item Both predictors excel at identifying correct rankings for post-RL performance, achieving $\geq 0.90$ Spearman correlation (30\% improvements); generalization loss provides advantageous prediction accuracy ($R^2$) for post-RL performance with up to 2x improvements.
\end{itemize}
\end{tcolorbox}\label{box:box1}

% \paragraph{How to use:}

% if only to rank,
% generalzation loss could rule out 
% Pass@k can rank 

% if desired to predict final performance,
% such as informing the trade-off between SFT costs and final perf,
% train a few model, fit a linear predictor, and predict on all models

\subsection{Use Case 2: Instance-Level Prediction}
This use case addresses the challenge of SFT data selection, an instance-level optimization problem. Here, the training pipeline is fixed, but we aim to select the optimal SFT dataset from a pool of candidates curated with different strategies (e.g., selecting for shortest/longest solutions, diversity, etc. \citet{ye2025limo}). This scenario tests whether strong SFT performance on a given dataset translates to a good final outcome after RL.

\begin{table}[h!]\vspace{-0em}
\centering
\caption{\small{Spearman's rank correlation between performance predicted from post-SFT models and the actual performance after RL. Pass@64 achieve notable margins over prediction from Pass@1.}\normalsize}\vspace{-0.0em}
\resizebox{0.8\linewidth}{!}{
\begin{tabular}{l|c|cc|c}
\toprule
\textbf{Spearman's Rank/} & Prediction based on SFT Pass@1 & Prediction based on  \\
\textbf{Correlation / Models} &  (avg. of 64) \textcolor{Orange}{\textbf{baseline}} &  SFT Pass@Large k (k=64) \\
\midrule
Llama3-8B-Instruct                        & 0.69 & \textbf{0.94} (\textcolor{Green}{+0.25})\\

Mistral-NeMo-12B-Instruct                 & 0.70 & \textbf{0.98} (\textcolor{Green}{+0.28})\\
% % Mistral-12B-Instruct                        & 0.71  & 0.82 &  \textbf{0.88} (\textcolor{Green}{+0.17})                  & 0.85                \\ 
% \midrule
% Large Budget ($5\sim20$B tokens)     & 46.56 (-1.48)          & 27.86 (-0.25)          \\  
\bottomrule
\end{tabular}
}
\end{table}\vspace{-0.0em}
\begin{table}[h!]
\centering
\caption{\small{Measuring prediction accuracy with coefficient of determination ($R^2$). We randomly select 50\% SFT models and fit a linear function between their post-SFT performance and performance after RL, and use it to predict for the other 50\% SFT models. We repeat random sampling for 100 times and report standard errors.}\normalsize}\vspace{-0.0em}
\resizebox{0.8\linewidth}{!}{
\begin{tabular}{l|c|cc|c}
\toprule
\textbf{Coefficient of determination} & Prediction based on SFT Pass@1 & Prediction based on  \\
\textbf{ ($R^2$) / Models} &  (avg. of 64) \textcolor{Orange}{\textbf{baseline}} &  SFT Pass@Large k (k=64) \\
\midrule
Llama3-8B-Instruct                        & 0.58 \tiny$\pm$ 0.20                 & \textbf{0.92} \tiny$\pm$ 0.05 \normalsize(\textcolor{Green}{+0.34})                \\

Mistral-NeMo-12B-Instruct                 & 0.73 \tiny$\pm$ 0.16                 & \textbf{0.98} \tiny$\pm$ 0.01 \normalsize(\textcolor{Green}{+0.25})                \\
% % Mistral-12B-Instruct                        & 0.71  & 0.82 &  \textbf{0.88} (\textcolor{Green}{+0.17})                  & 0.85                \\ 
% \midrule
% Large Budget ($5\sim20$B tokens)     & 46.56 (-1.48)          & 27.86 (-0.25)          \\  
\bottomrule
\end{tabular}
}\vspace{-0em}
\end{table}

\textit{In this scenario, the generalization loss predictor is not applicable.} Since each SFT dataset comes from a different distribution, the validation loss includes a distributional gap component in addition to generalization error. Without a common, representative validation set, it is difficult to make a fair comparison. Pass@large k metric proves to be exceptionally robust. Since it measures the model's inherent capability to produce correct solutions, it is less sensitive to distributional shifts in the training data. It can be used to effectively rank different SFT datasets and select the one with the highest potential for RL, without needing to run any RL experiments for calibration.

\begin{tcolorbox}[title=\text{Takeaway 2: Instance-level Prediction},colback=lightgray!30!white,colframe=black]
\begin{itemize}[leftmargin=*]
    \item Pass@large k turns out highly accurate and robust in instance-level predictions, improving Spearman correlation by up to 36\% and prediction accuracy ($R^2$)by up to 59\%. It effectively identifies datasets for strong post-RL performance and predicts RL outcomes.
\item Generalization loss is not applicable for instance-level selection due to distributional gaps between different datasets. 
\end{itemize}
\end{tcolorbox}\label{box:box2}\vspace{-0.0em}

\paragraph{How to use them in practice?} Our metrics support two primary workflows. If the goal is simply to \textit{rank} SFT candidates, one can use generalization loss to quickly filter out clearly suboptimal models (i.e., those with both low performance and high loss). Then, Pass@large k can be used to reliably rank the remaining candidates to identify the most promising one. If the goal is to \textit{predict the final performance value}—for instance, to inform trade-offs between SFT costs and expected gains—practitioners can run RL on a small number of SFT models to gather calibration data. A linear predictor can then be fitted using our proposed metrics, allowing for accurate performance estimation across all SFT candidates without the need for exhaustive RL runs.

\vspace{-0.5em}\section{Conclusions}\vspace{-0.5em}
This work confronts a critical quagmire in reasoning post-training: the common assumption that high SFT scores guarantee strong performance after subsequent RL. Through extensive experimentation with Llama3/Mistral-Nemo/Qwen3 models spending $>$1M GPU hours, we provide broad counter-examples where SFT performance is often misleading or biased toward simpler/repeated data. Our primary contribution is the identification and validation of two more reliable predictors for post-RL success: generalization loss on held-out reasoning examples and Pass@large k accuracy, improving prediction accuracy ($R^2$) and Spearman's rank correlation by up to 0.5 (2x) over prediction from post-SFT performance. By allowing practitioners to better predict the final outcome, our work helps de-risk the expensive RL stage and streamline the entire post-training pipeline. We will open-source our enhanced evaluation tool to facilitate broader adoption and further research.
This work focuses on mathematical reasoning. A natural \textbf{next step} is to study the topic in a wider range of reasoning tasks (e.g., coding, science) and agentic use cases; Our study is limited to the prevailing paradigm of online RL with GRPO. The relationship between SFT characteristics and post-RL performance with other methods such as offline RL/DPO or other RL algorithms may worth further explorations; Directly evaluating Pass@large k requires repeating evaluation for at least k times, which becomes computational expensive with long sequence lengths. Estimating Pass@k accuracy from that of smaller k holds the promise for more efficient evaluations \citep{schaeffer2025large}.

\section*{Acknowledgment}
Ruoxi Jia and the ReDS lab acknowledge support from the National Science Foundation through grants IIS-2312794, IIS-2313130, OAC-2239622.

% \subsubsection*{Acknowledgments}
% Use unnumbered third level headings for the acknowledgments. All
% acknowledgments, including those to funding agencies, go at the end of the paper.

\clearpage
\bibliographystyle{assets/plainnat}
\bibliography{paper}

\clearpage
\appendix

\begin{appendices}

\startcontents[appendices]
% \appendixpage
\printcontents[appendices]{}{1}{\setcounter{tocdepth}{2}}

\clearpage

\section{Additional SFT-RL Examples and Visualizations}\label{app:vis}
\subsection{Llama3-8B-Instruct}
\begin{figure}[h!]
    \centering
        \includegraphics[width=\textwidth]{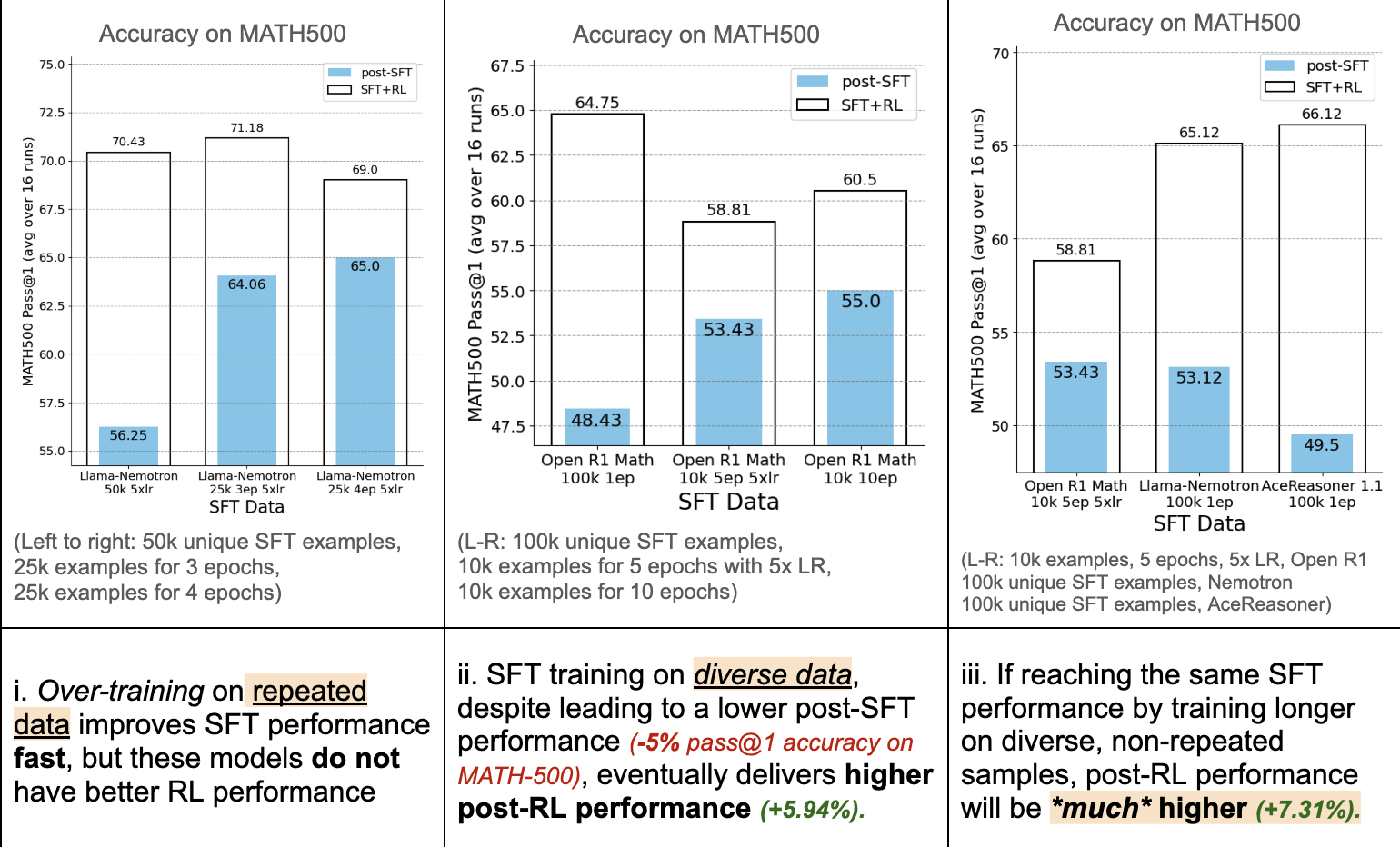}
        \vspace{-0em}
    \caption{\small{Llama3-8B-Instruct undergone SFT-RL with SFT examples from Llama-Nemotron-SFT/AceReasoner1.1-SFT/OpenR1-Math dataset and RLVR via GRPO on MATH dataset (train-split). Reporting average Pass@1 performance on MATH-500 (test-split). \textit{High SFT scores can be biased toward simpler or more homogeneous data and are not reliably predictive of subsequent RL gains or post-training effectiveness.} SFT on fewer unique examples repeated for more training epochs (ep) or/and with a larger learning rate (LR) leads to higher accuracy on reasoning benchmarks such as MATH-500 (+8.75\% vs. non-repeated data, left figure). However, models trained this way show smaller improvements during RL (-1.43\% vs. non-repeated). In contrast, SFT on more diverse, non-repeated data—despite yielding lower initial SFT performance (-5\% vs. repeated data, middle/right figure)—results in significantly better post-RL performance (+5.94\%). 
}}\normalsize\vspace{-0em}\label{fig:app-l8}
\end{figure}

Figure \ref{fig:app-l8} shows results on Llama3-8B-Instruct undergone SFT-RL with SFT examples from Llama-Nemotron-SFT/AceReasoner1.1-SFT/OpenR1-Math \citep{openr1} dataset and RLVR via GRPO on MATH dataset (train-split). Reporting average Pass@1 performance on MATH-500 (test-split). \textit{High SFT scores can be biased toward simpler or more homogeneous data and are not reliably predictive of subsequent RL gains or post-training effectiveness.} SFT on fewer unique examples repeated for more training epochs (ep) or/and with a larger learning rate (LR) leads to higher accuracy on reasoning benchmarks such as MATH-500 (+8.75\% vs. non-repeated data, left figure). However, models trained this way show smaller improvements during RL (-1.43\% vs. non-repeated). In contrast, SFT on more diverse, non-repeated data—despite yielding lower initial SFT performance (-5\% vs. repeated data, middle/right figure)—results in significantly better post-RL performance (+5.94\%). 

\begin{figure}[h!]
    \centering
        \includegraphics[width=\textwidth]{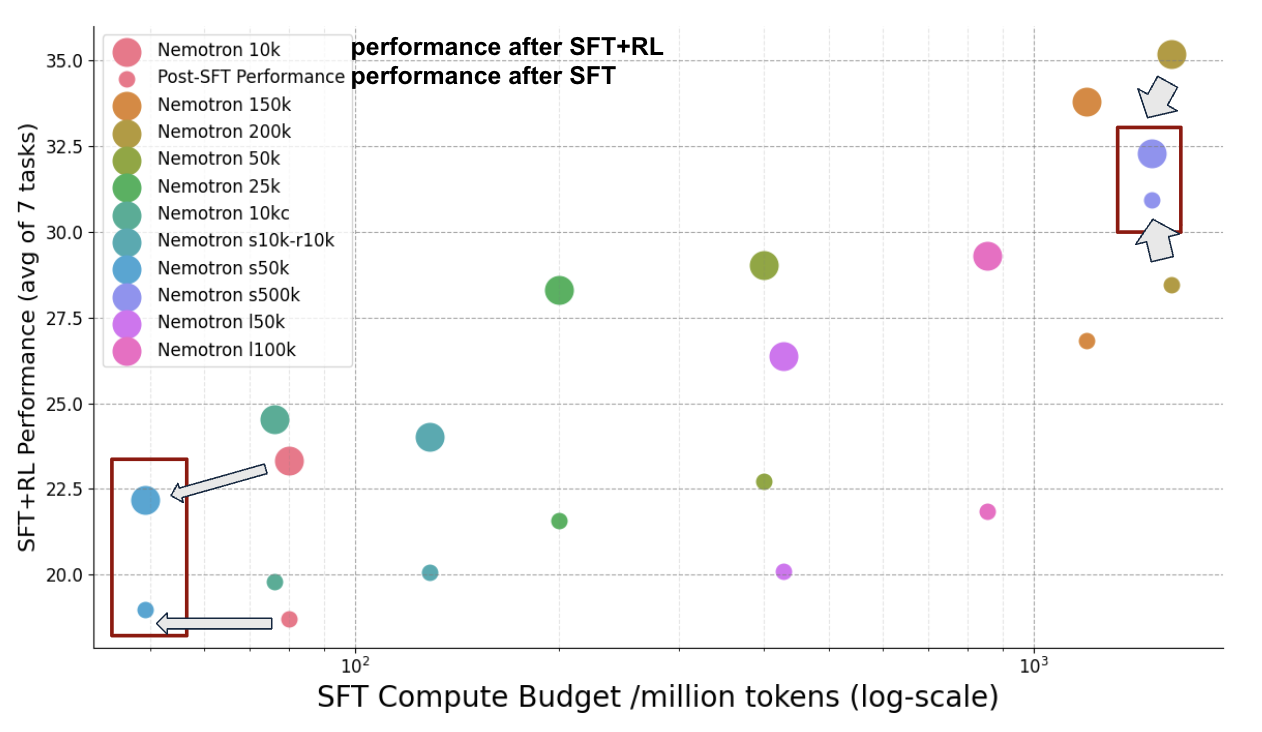}
        \vspace{-0em}
    \caption{\small{Llama3-8B-Instruct undergone SFT-RL with SFT examples from Llama-Nemotron-SFT dataset and RLVR via GRPO on MATH dataset (train-split). Reporting Pass@1 performance averaged over 7 math benchmarks. High SFT scores can be biased toward \textit{simpler examples} and are not reliably predictive of subsequent RL gains or scaled-up post-training effectiveness. For example, training on shortest examples (e.g., s10k, s500k) led to faster performance improvements than training on randomly sampled examples (e.g., 10k, 200k) during SFT (lower smaller dots). These shorter examples are closer to the model’s original generations and easier to learn, though, these are not best examples for the model to gain reasoning capabilities in preparation for RL. The final performance after RL (upper larger dots) is significantly lower. }\normalsize}\label{fig:app-l8-compute}\vspace{-0em}
\end{figure}

Figure \ref{fig:app-l8-compute} shows results on Llama3-8B-Instruct undergone SFT-RL with SFT examples from Llama-Nemotron-SFT dataset and RLVR via GRPO on MATH dataset (train-split). Reporting Pass@1 performance averaged over 7 math benchmarks. High SFT scores can be biased toward \textit{simpler examples} and are not reliably predictive of subsequent RL gains or scaled-up post-training effectiveness. For example, training on shortest examples (e.g., s10k, s500k) led to faster performance improvements than training on randomly sampled examples (e.g., 10k, 200k) during SFT (lower smaller dots). These shorter examples are closer to the model’s original generations and easier to learn, though, these are not best examples for the model to gain reasoning capabilities in preparation for RL. The final performance after RL (upper larger dots) is significantly lower.

\subsection{Mistral-NeMo-12B-Instruct}

\begin{figure}[h!]
    \centering
        \includegraphics[width=\textwidth]{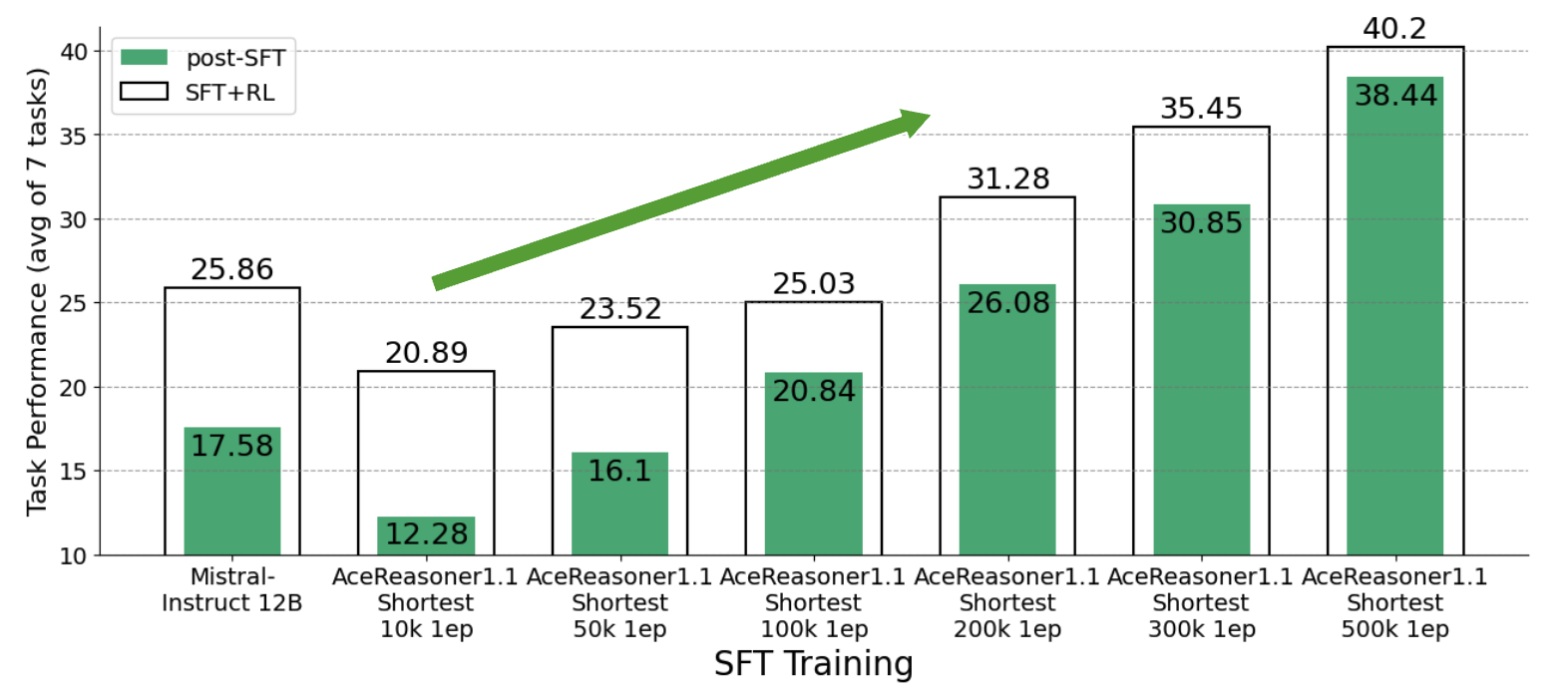}
        \vspace{-0em}
    \caption{\small{Mistral-NeMo-12B-Instruct undergone SFT-RL with \textbf{shortest} SFT examples from AceReasoner1.1-SFT dataset and RLVR via GRPO on DeepScaleR dataset. Reporting Pass@1 performance averaged over 7 math benchmarks. With increasing SFT examples, Mistral's post-SFT performance first dips and then gradually recovers and improves to performance better than before SFT training. Compared to the base model, the final performance after RL also first dips and then gradually goes up and improves to a better level. \textit{Notably, post-RL performance recovers to the same level as the base model slower than the post-SFT performance.} The post-SFT and post-RL performance trends are not identical.} \normalsize}\label{fig:app-mis}\vspace{-0em}
\end{figure}

Figure \ref{fig:app-mis} shows results on Mistral-NeMo-12B-Instruct undergone SFT-RL with \textbf{shortest} SFT examples from AceReasoner1.1-SFT dataset and RLVR via GRPO on DeepScaleR dataset. Reporting Pass@1 performance averaged over 7 math benchmarks. With increasing SFT examples, Mistral's post-SFT performance first dips and then gradually recovers and improves to performance better than before SFT training. Compared to the base model, the final performance after RL also first dips and then gradually goes up and improves to a better level. \textit{Notably, post-RL performance recovers to the same level as the base model slower than the post-SFT performance.} The post-SFT and post-RL performance trends are not identical.

\subsection{Qwen3-4B-base}
\begin{figure}[h!]
    \centering
        \includegraphics[width=\textwidth]{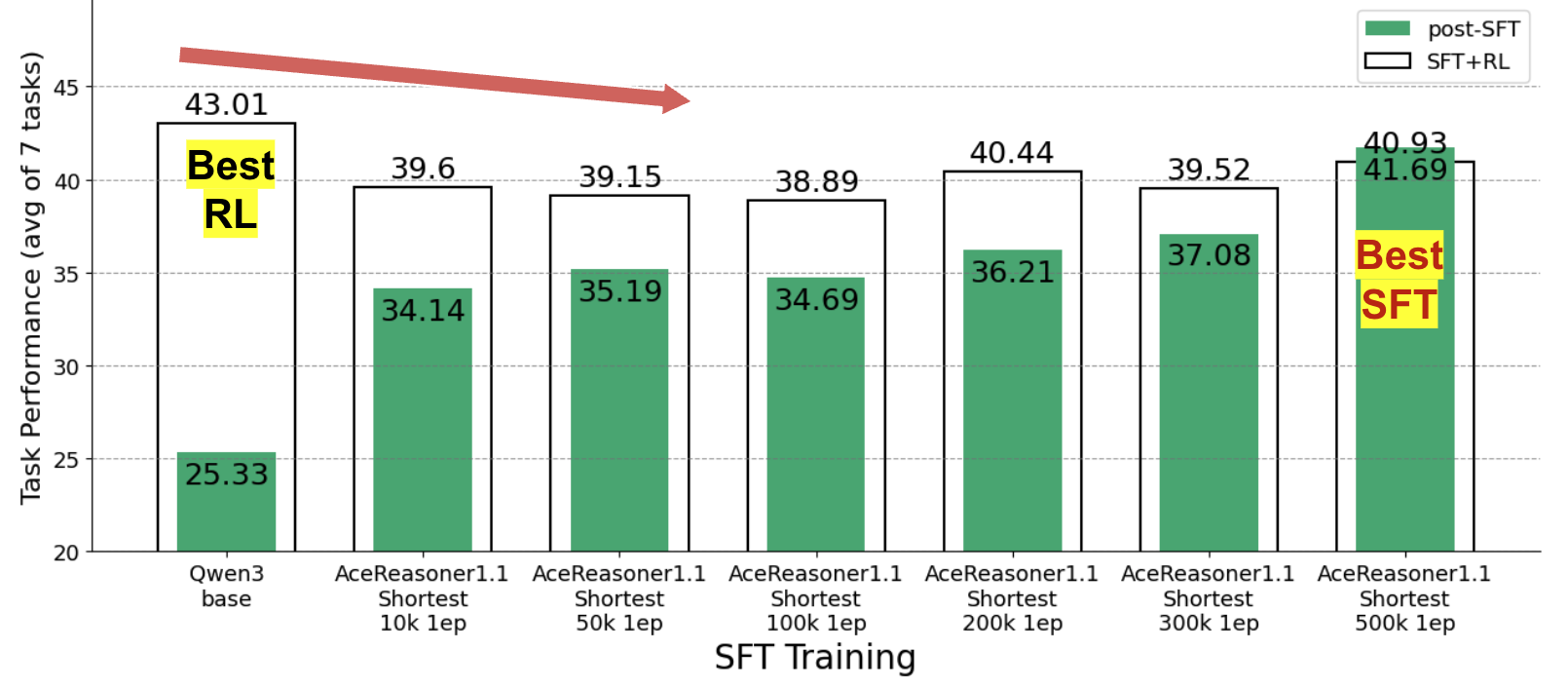}
        \vspace{-0em}
    \caption{\small{Qwen3-4B-base undergone SFT-RL with \textbf{shortest} SFT examples from AceReasoner1.1-SFT dataset and RLVR via GRPO on DeepScaleR dataset. Reporting Pass@1 performance averaged over 7 math benchmarks. With increasing SFT examples, Qwen3's post-SFT performances appear uncorrelated with the final performance after RL, where the latter remains the same despite the substantially improved SFT performance.} \normalsize}\vspace{-0em}\label{fig:app-q3-s}
\end{figure}

Figure \ref{fig:app-q3-s} shows results on Qwen3-4B-base undergone SFT-RL with \textbf{shortest} SFT examples from AceReasoner1.1-SFT dataset and RLVR via GRPO on DeepScaleR dataset. Reporting Pass@1 performance averaged over 7 math benchmarks. With increasing SFT examples, Qwen3's post-SFT performances appear uncorrelated with the final performance after RL, where the latter remains the same despite the substantially improved SFT performance.

\begin{figure}[h!]
    \centering
        \includegraphics[width=0.7\textwidth]{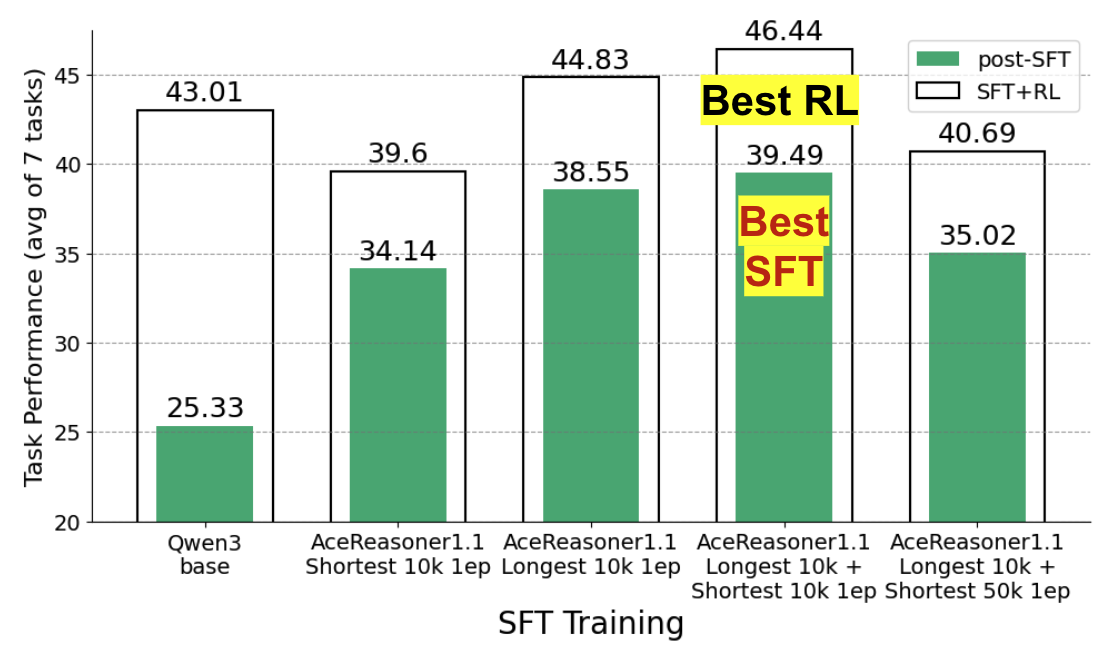}
        \vspace{-0em}
    \caption{\small{Qwen3-4B-base undergone SFT-RL with \textbf{shortest/Longest/Longest+Shortest} SFT examples from AceReasoner1.1-SFT dataset and RLVR via GRPO on DeepScaleR dataset. Reporting Pass@1 performance averaged over 7 math benchmarks. All SFT training substantially improves Qwen3's post-SFT performance, but the final performance after RL is mixed. Training on \textbf{Longest} and 10k \textbf{Longest}+10k \textbf{Shortest} SFT examples lead to visibly improved final performance after RL where the latter achieves the best final performance for Qwen3 models in this work. Other SFT training lead to significantly degraded final performance after RL.}\normalsize} \vspace{-0em}\label{fig:app-q3-d}
\end{figure}

Figure \ref{fig:app-q3-d} shows results on Qwen3-4B-base undergone SFT-RL with \textbf{Shortest/Longest/Longest+Shortest} SFT examples from AceReasoner1.1-SFT dataset and RLVR via GRPO on DeepScaleR dataset. Reporting Pass@1 performance averaged over 7 math benchmarks. All SFT training substantially improves Qwen3's post-SFT performance, but the final performance after RL is mixed. Training on \textbf{Longest} and 10k \textbf{Longest}+10k \textbf{Shortest} SFT examples lead to visibly improved final performance after RL where the latter achieves the best final performance for Qwen3 models in this work. Other SFT training lead to significantly degraded final performance after RL.

\section{Implementation Details}\label{app:setup}
All experiments (SFT, RL, evaluation) are conducted on individual AWS \citep{mathew2014overview} node with 8x NVIDIA A100 80GB GPU. 
 Experiments spent $>$1M GPU hours on NVIDIA A100 80GB. 
We repeat RL training for 4+ runs on each data recipe and training paradigm (each run takes up to 5 days), conduct 4+ evaluations on different checkpoints across RL training run, and report the best performance for the model. We set the max sequence length to 8k tokens throughout SFT, RL, and evaluation.

\subsection{Models and Datasets}
We conduct three sets of experiments with SFT-RL post-training. On \textbf{Llama3-8B-Instruct} models, we conduct SFT training with examples from Llama-Nemotron dataset (where we only select math samples with responses generated by QwQ-32B \citep{qwq32b} or DeepSeek R1 \citep{guo2025deepseek}, hereinafter the same) and RL training on MATH dataset (train-split) \citep{hendrycks2021measuring}; on \textbf{Mistral-Nemo-12B-Instruct} and Qwen3-4B-base,  we conduct SFT training with examples from AceReasoner1.1-SFT dataset and RL training on DeepScaleR dataset \citep{deepscaler2025}. For all models, we conduct RL training for 3 epochs where each run  takes up to 5 days.

\subsection{Training}
We conduct SFT training with LLaMA-Factory \citep{zheng2024llamafactory} using learning rates lr=1e-5 and global batch size = 128, and RL training via GRPO with verl \citep{sheng2024hybridflow} using learning rates lr=1e-6 and global batch size = 128. We sample 16 rollouts for each question with temperature=1.0. We set KL loss coefficient=0 and entropy coefficient=0.001.  

\subsection{Evaluation}
Evaluations are conducted with pipelines originally developed in this work based on vllm \citep{kwon2023efficient} and HuggingFace’s math-verify \citep{huggingface2025math}, enabling efficient inference with performant and accurate verification. We ran evaluations with the same template and generation configuration as in RL, using decoding temperature t=1.0 and the standard reasoning prompt (\texttt{“Let's think step by step and output the final answer within \textbackslash\textbackslash boxed\{\}.”}). 

We evaluate task performance on 7 math benchmarks, including \textbf{MATH-500} \citep{hendrycks2021measuring}, \textbf{AIME 1983-2024} \citep{aime_1983_2024}, \textbf{GSM8k} \citep{cobbe2021gsm8k}, \textbf{AIME 2025} \citep{aime_2025}, \textbf{AMC} \citep{amc}, \textbf{Olympiad} \citep{he2024olympiadbench}, \textbf{Minerva} \citep{lewkowycz2022solving}, and report model performance as Pass@1 averaged over 64 repetitions and across 7 tasks. For the proposed predictors, we evaluate the generalization loss on the validation set of the SFT data and Pass@64 accuracy averaged over 256 repetitions.

\section{Additional Experimental Results}\label{app:res}

\subsection{Dataset-level}
Table \ref{app:table-d-l8-r2} shows results on Llama3-8B-Instruct undergone SFT-RL with SFT examples from Llama-Nemotron-SFT dataset and RLVR via GRPO on MATH dataset (train-split). Reporting Pass@1 performance averaged over 7 math benchmarks. Measuring prediction accuracy with coefficient of determination ($R^2$) varying the ratio of fit-validation datapoints. We randomly select x SFT models and fit a linear function between their post-SFT performance and performance after RL, and use it to predict for the rest SFT models. We repeat random sampling for 100 times and report standard errors.

Table \ref{app:table-d-m-r2} shows results on Mistral-NeMo-12B-Instruct undergone SFT-RL with SFT examples from AceReasoner1.1-SFT dataset and RLVR via GRPO on DeepScaleR dataset.  Measuring prediction accuracy with coefficient of determination ($R^2$) varying the ratio of fit-validation datapoints. Reporting Pass@1 performance averaged over 7 math benchmarks. We randomly select x SFT models and fit a linear function between their post-SFT performance and performance after RL, and use it to predict for the rest SFT models. We repeat random sampling for 100 times and report standard errors.

\begin{table}[h!]
\centering
\caption{Llama3-8B-Instruct undergone SFT-RL with SFT examples from Llama-Nemotron-SFT dataset and RLVR via GRPO on MATH dataset (train-split). Reporting Pass@1 performance averaged over 7 math benchmarks. Measuring prediction accuracy with coefficient of determination ($R^2$) varying the ratio of fit-validation datapoints. We randomly select x SFT models and fit a linear function between their post-SFT performance and performance after RL, and use it to predict for the rest SFT models. We repeat random sampling for 100 times and report standard errors.}\label{app:table-d-l8-r2}
\resizebox{1.0\linewidth}{!}{
\begin{tabular}{l|c|cc|c}
\toprule
\textbf{No. of Fitting-Validation } & Prediction based on & Prediction based on & Prediction based on & Avg. Prediction from \\
\textbf{Datapoints / Coefficient of } & SFT Pass@1 (avg. &  SFT Generalization&  SFT Pass@Large k & SFT Gen. Loss + \\
\textbf{determination ($R^2$)} & of 64) \textcolor{Orange}{\textbf{baseline}}&  Loss &   (k=64) &  Pass@Large k (64)\\
\midrule

Fitting: 3; Validation: 13                        & 0.48 \tiny$\pm$ 0.40       & 0.80 \tiny$\pm$ 0.33 & 0.80 \tiny$\pm$ 0.23          & \textbf{0.86} \tiny$\pm$ 0.22 \normalsize(\textcolor{Green}{+0.38})                \\ 

Fitting: 4; Validation: 12                        & 0.57 \tiny$\pm$ 0.29       & 0.82 \tiny$\pm$ 0.21 & 0.84 \tiny$\pm$ 0.15          & \textbf{0.92} \tiny$\pm$ 0.08 \normalsize(\textcolor{Green}{+0.35})                \\ 

Fitting: 5; Validation: 11                        & 0.57 \tiny$\pm$ 0.29       & 0.88 \tiny$\pm$ 0.09 & 0.87 \tiny$\pm$ 0.10          & \textbf{0.94} \tiny$\pm$ 0.04 \normalsize(\textcolor{Green}{+0.37})                \\ 

Fitting: 6; Validation: 10                        & 0.57 \tiny$\pm$ 0.26       & 0.89 \tiny$\pm$ 0.07 & 0.87 \tiny$\pm$ 0.10          & \textbf{0.95} \tiny$\pm$ 0.03 \normalsize(\textcolor{Green}{+0.38})                \\ 

Fitting: 7; Validation: 9                        & 0.64 \tiny$\pm$ 0.19       & 0.89 \tiny$\pm$ 0.07 & 0.90 \tiny$\pm$ 0.06          & \textbf{0.95} \tiny$\pm$ 0.05 \normalsize(\textcolor{Green}{+0.31})                \\ 

Fitting: 8; Validation: 8                       & 0.64 \tiny$\pm$ 0.20       & 0.88 \tiny$\pm$ 0.08 & 0.88 \tiny$\pm$ 0.08          & \textbf{0.93} \tiny$\pm$ 0.05 \normalsize(\textcolor{Green}{+0.29})                \\ 

Fitting: 10; Validation: 6                        & 0.59 \tiny$\pm$ 0.33       & 0.85 \tiny$\pm$ 0.17 & 0.85 \tiny$\pm$ 0.15          & \textbf{0.94} \tiny$\pm$ 0.05 \normalsize(\textcolor{Green}{+0.35})                \\ 

Fitting: 12; Validation: 4                       & 0.54 \tiny$\pm$ 0.43       & 0.86 \tiny$\pm$ 0.18 & 0.81 \tiny$\pm$ 0.23          & \textbf{0.91} \tiny$\pm$ 0.12 \normalsize(\textcolor{Green}{+0.37})                \\ 
\bottomrule
\end{tabular}
}
\end{table}

\begin{table}[h!]
\centering
\caption{Mistral-NeMo-12B-Instruct undergone SFT-RL with SFT examples from AceReasoner1.1-SFT dataset and RLVR via GRPO on DeepScaleR dataset.  Measuring prediction accuracy with coefficient of determination ($R^2$) varying the ratio of fit-validation datapoints. Reporting Pass@1 performance averaged over 7 math benchmarks. We randomly select x SFT models and fit a linear function between their post-SFT performance and performance after RL, and use it to predict for the rest SFT models. We repeat random sampling for 100 times and report standard errors.}\label{app:table-d-m-r2}
\resizebox{1.0\linewidth}{!}{
\begin{tabular}{l|c|cc|c}
\toprule
\textbf{No. of Fitting-Validation } & Prediction based on & Prediction based on & Prediction based on & Avg. Prediction from \\
\textbf{Datapoints / Coefficient of } & SFT Pass@1 (avg. &  SFT Generalization&  SFT Pass@Large k & SFT Gen. Loss + \\
\textbf{determination ($R^2$)} & of 64) \textcolor{Orange}{\textbf{baseline}}&  Loss &   (k=64) &  Pass@Large k (64)\\
\midrule

Fitting: 3; Validation: 7                        & 0.32 \tiny$\pm$ 0.39       & \textbf{0.73}  \tiny$\pm$ 0.41 \normalsize(\textcolor{Green}{+0.41}) & 0.52 \tiny$\pm$ 0.31          & 0.61 \tiny$\pm$ 0.38               \\
Fitting: 4; Validation: 6                        & 0.27 \tiny$\pm$ 0.36       & \textbf{0.75} \tiny$\pm$ 0.34 \normalsize(\textcolor{Green}{+0.48}) & 0.51 \tiny$\pm$ 0.37          & 0.69 \tiny$\pm$ 0.26               \\
Fitting: 5; Validation: 5                        & 0.29 \tiny$\pm$ 0.38       & \textbf{0.79} \tiny$\pm$ 0.26 \normalsize(\textcolor{Green}{+0.50})  & 0.57 \tiny$\pm$ 0.32          & 0.72 \tiny$\pm$ 0.24              \\
Fitting: 6; Validation: 4                        & 0.37 \tiny$\pm$ 0.37       & \textbf{0.78} \tiny$\pm$ 0.25 \normalsize(\textcolor{Green}{+0.41}) & 0.57 \tiny$\pm$ 0.37          & 0.67 \tiny$\pm$ 0.35               \\
Fitting: 7; Validation: 3                        & 0.36 \tiny$\pm$ 0.36       & \textbf{0.77} \tiny$\pm$ 0.30 \normalsize(\textcolor{Green}{+0.41})  & 0.57 \tiny$\pm$ 0.35          & 0.66 \tiny$\pm$ 0.37              \\
Fitting: 8; Validation: 2                        & 0.31 \tiny$\pm$ 0.46       & \textbf{0.68} \tiny$\pm$ 0.36  \normalsize(\textcolor{Green}{+0.37}) & 0.47 \tiny$\pm$ 0.54          & 0.64 \tiny$\pm$ 0.37              \\
\bottomrule
\end{tabular}
}
\end{table}

\subsection{Instance-level}

Table \ref{app:table-i-l8-r2} shows results on Llama3-8B-Instruct undergone SFT-RL with SFT examples from Llama-Nemotron-SFT dataset and RLVR via GRPO on MATH dataset (train-split). Reporting Pass@1 performance averaged over 7 math benchmarks. Measuring prediction accuracy with coefficient of determination ($R^2$) varying the ratio of fit-validation datapoints. We randomly select x SFT models and fit a linear function between their post-SFT performance and performance after RL, and use it to predict for the rest SFT models. We repeat random sampling for 100 times and report standard errors.

Table \ref{app:table-i-l8-s} shows results on Llama3-8B-Instruct undergone SFT-RL with SFT examples from Llama-Nemotron-SFT dataset and RLVR via GRPO on MATH dataset (train-split). Reporting Pass@1 performance averaged over 7 math benchmarks. Spearman's rank correlation between performance predicted from post-SFT models and the actual performance after RL, grouped by different SFT training budget.

\begin{table}[h!]
\centering
\caption{Llama3-8B-Instruct undergone SFT-RL with SFT examples from Llama-Nemotron-SFT dataset and RLVR via GRPO on MATH dataset (train-split). Reporting Pass@1 performance averaged over 7 math benchmarks. Measuring prediction accuracy with coefficient of determination ($R^2$) varying the ratio of fit-validation datapoints. We randomly select x SFT models and fit a linear function between their post-SFT performance and performance after RL, and use it to predict for the rest SFT models. We repeat random sampling for 100 times and report standard errors.}\label{app:table-i-l8-r2}
\resizebox{0.90\linewidth}{!}{
\begin{tabular}{l|cc}
\toprule
\textbf{No. of Fitting-Validation Datapoints/} & Prediction based on SFT Pass@1 & Prediction based on  \\
\textbf{Coefficient of determination ($R^2$)} &  (avg. of 64) \textcolor{Orange}{\textbf{baseline}} &  SFT Pass@Large k (k=64) \\
\midrule
Fitting: 3; Validation: 14                        & 0.40 \tiny$\pm$ 0.31                 & \textbf{0.89} \tiny$\pm$ 0.10 \normalsize(\textcolor{Green}{+0.49})                \\ 
Fitting: 4; Validation: 13                        & 0.49 \tiny$\pm$ 0.30                 & \textbf{0.89} \tiny$\pm$ 0.17 \normalsize(\textcolor{Green}{+0.40})                \\ 
Fitting: 5; Validation: 12                        & 0.55 \tiny$\pm$ 0.22                 & \textbf{0.91} \tiny$\pm$ 0.05 \normalsize(\textcolor{Green}{+0.36})                \\
Fitting: 6; Validation: 11                        & 0.54 \tiny$\pm$ 0.30                 & \textbf{0.92} \tiny$\pm$ 0.04 \normalsize(\textcolor{Green}{+0.38})                \\
Fitting: 7; Validation: 10                        & 0.55 \tiny$\pm$ 0.24                 & \textbf{0.92} \tiny$\pm$ 0.04 \normalsize(\textcolor{Green}{+0.37})                \\
Fitting: 8; Validation: 9                        & 0.58 \tiny$\pm$ 0.20                 & \textbf{0.92} \tiny$\pm$ 0.05 \normalsize(\textcolor{Green}{+0.34})                \\
Fitting: 10; Validation: 7                        & 0.56 \tiny$\pm$ 0.25                 & \textbf{0.92} \tiny$\pm$ 0.05 \normalsize(\textcolor{Green}{+0.36})                \\
Fitting: 12; Validation: 5                        & 0.57 \tiny$\pm$ 0.28                 & \textbf{0.92} \tiny$\pm$ 0.05 \normalsize(\textcolor{Green}{+0.35})                \\
\bottomrule
\end{tabular}
}
\end{table}

\begin{table}[h!]
\centering
\caption{Llama3-8B-Instruct undergone SFT-RL with SFT examples from Llama-Nemotron-SFT dataset and RLVR via GRPO on MATH dataset (train-split). Reporting Pass@1 performance averaged over 7 math benchmarks. Spearman's rank correlation between performance predicted from post-SFT models and the actual performance after RL, grouped by different SFT training budget.}\label{app:table-i-l8-s}
\resizebox{0.85\linewidth}{!}{
\begin{tabular}{l|cc}
\toprule
\textbf{SFT Compute Budget/} & Prediction based on SFT Pass@1 & Prediction based on  \\
\textbf{Spearman's Rank Correlation} &  (avg. of 64) \textcolor{Orange}{\textbf{baseline}} &  SFT Pass@Large k (k=64) \\
\midrule
Low Budget ($<2$B tokens)                        & 0.77                  & \textbf{0.99} (\textcolor{Green}{+0.22})                 \\ 
Medium Budget ($2\sim5$B tokens)     & 0.60          & \textbf{0.90} (\textcolor{Green}{+0.30})         \\ 
High Budget ($5\sim20$B tokens)     & 0.70          & \textbf{0.94} (\textcolor{Green}{+0.24})         \\  
\midrule
Average & 0.69 & \textbf{0.94} (\textcolor{Green}{+0.25})\\
% \midrule
% Large Budget ($5\sim20$B tokens)     & 46.56 (-1.48)          & 27.86 (-0.25)          \\  
\bottomrule
\end{tabular}
}
\end{table}

Table \ref{app:table-i-m-r2} shows results on Mistral-NeMo-12B-Instruct undergone SFT-RL with SFT examples from AceReasoner1.1-SFT dataset and RLVR via GRPO on DeepScaleR dataset.  Measuring prediction accuracy with coefficient of determination ($R^2$) varying the ratio of fit-validation datapoints. Mistral-NeMo-12B-Instruct undergone SFT-RL with SFT examples from AceReasoner1.1-SFT dataset and RLVR via GRPO on DeepScaleR dataset. Reporting Pass@1 performance averaged over 7 math benchmarks. We randomly select x SFT models and fit a linear function between their post-SFT performance and performance after RL, and use it to predict for the rest SFT models. We repeat random sampling for 100 times and report standard errors.
\begin{table}[h!]
\centering
\caption{Mistral-NeMo-12B-Instruct undergone SFT-RL with SFT examples from AceReasoner1.1-SFT dataset and RLVR via GRPO on DeepScaleR dataset.  Measuring prediction accuracy with coefficient of determination ($R^2$) varying the ratio of fit-validation datapoints. Mistral-NeMo-12B-Instruct undergone SFT-RL with SFT examples from AceReasoner1.1-SFT dataset and RLVR via GRPO on DeepScaleR dataset. Reporting Pass@1 performance averaged over 7 math benchmarks. We randomly select x SFT models and fit a linear function between their post-SFT performance and performance after RL, and use it to predict for the rest SFT models. We repeat random sampling for 100 times and report standard errors.}\label{app:table-i-m-r2}
\resizebox{0.90\linewidth}{!}{
\begin{tabular}{l|cc}
\toprule
\textbf{No. of Fitting-Validation Datapoints/} & Prediction based on SFT Pass@1 & Prediction based on  \\
\textbf{Coefficient of determination ($R^2$)} &  (avg. of 64) \textcolor{Orange}{\textbf{baseline}} &  SFT Pass@Large k (k=64) \\
\midrule
Fitting: 2; Validation: 10                        & 0.55 \tiny$\pm$ 0.42                 & \textbf{0.87} \tiny$\pm$ 0.29 \normalsize(\textcolor{Green}{+0.32})                \\ 
Fitting: 3; Validation: 9                        & 0.71 \tiny$\pm$ 0.15                 & \textbf{0.94} \tiny$\pm$ 0.18 \normalsize(\textcolor{Green}{+0.23})                \\ 
Fitting: 4; Validation: 8                        & 0.69 \tiny$\pm$ 0.22                 & \textbf{0.98} \tiny$\pm$ 0.03 \normalsize(\textcolor{Green}{+0.29})                \\
Fitting: 5; Validation: 7                        & 0.75 \tiny$\pm$ 0.10                 & \textbf{0.98} \tiny$\pm$ 0.01 \normalsize(\textcolor{Green}{+0.23})                \\
Fitting: 6; Validation: 6                        & 0.73 \tiny$\pm$ 0.16                 & \textbf{0.98} \tiny$\pm$ 0.01 \normalsize(\textcolor{Green}{+0.25})                \\
Fitting: 8; Validation: 4                        & 0.69 \tiny$\pm$ 0.35                 & \textbf{0.97} \tiny$\pm$ 0.03 \normalsize(\textcolor{Green}{+0.28})                \\
Fitting: 10; Validation: 2                        & 0.68 \tiny$\pm$ 0.42                 & \textbf{0.91} \tiny$\pm$ 0.17 \normalsize(\textcolor{Green}{+0.23})                \\
\bottomrule
\end{tabular}
}
\end{table}

Table \ref{app:table-i-m-s} shows results on Mistral-NeMo-12B-Instruct undergone SFT-RL with SFT examples from AceReasoner1.1-SFT dataset and RLVR via GRPO on DeepScaleR dataset. Reporting Pass@1 performance averaged over 7 math benchmarks. Spearman's rank correlation between performance predicted from post-SFT models and the actual performance after RL, grouped by different SFT training budget.
\begin{table}[h!]
\centering
\caption{Mistral-NeMo-12B-Instruct undergone SFT-RL with SFT examples from AceReasoner1.1-SFT dataset and RLVR via GRPO on DeepScaleR dataset. Reporting Pass@1 performance averaged over 7 math benchmarks. Spearman's rank correlation between performance predicted from post-SFT models and the actual performance after RL, grouped by different SFT training budget.}\label{app:table-i-m-s}
\resizebox{0.85\linewidth}{!}{
\begin{tabular}{l|cc}
\toprule
\textbf{SFT Compute Budget/} & Prediction based on SFT Pass@1 & Prediction based on  \\
\textbf{Spearman's Rank Correlation} &  (avg. of 64) \textcolor{Orange}{\textbf{baseline}} &  SFT Pass@Large k (k=64) \\
\midrule
Low Budget ($<2$B tokens)                        & 0.80                  & \textbf{0.95} (\textcolor{Green}{+0.25})                 \\ 
Medium Budget ($2\sim5$B tokens)     & 0.80          & \textbf{1.00} (\textcolor{Green}{+0.20})         \\ 
High Budget ($5\sim20$B tokens)     & 0.50          & \textbf{1.00} (\textcolor{Green}{+0.50})         \\  
\midrule
Average & 0.70 & \textbf{0.98} (\textcolor{Green}{+0.28})\\
% \midrule
% Large Budget ($5\sim20$B tokens)     & 46.56 (-1.48)          & 27.86 (-0.25)          \\  
\bottomrule
\end{tabular}
}
\end{table}

\clearpage
\section{Sample SFT Examples}\label{app:sample}
\subsection{Shortest Examples}
A sample Shortest Examples from Llama-Nemotron Post-Training Dataset \citep{singhal2025llama} \textbf{($\sim$750 tokens)}:

\textbf{$[$Input$]$}: \begin{lstlisting}[breaklines=true]
Solve the following math problem. Make sure to put the answer (and only answer) inside \\boxed{}.\n\nThe probability of snow on 1st January is 1/20. What is the probability that snow will fall on the next 1st January?"
\end{lstlisting}
\textbf{$[$Output$]$}: 
\begin{lstlisting}[breaklines=true]
<think> Okay, so I have this problem here: "The probability of snow on 1st January is 1/20. What is the probability that snow will fall on the next 1st January?" Hmm, let's see. Alright, the question is asking for the probability that it will snow next January 1st, given that the probability is 1/20. Wait, is there more to this? Let me check again.  Wait, maybe I misread. The original probability is 1/20 for snow on January 1st, and they want the probability it will snow on the next January 1st. Hmm. That seems straightforward. But maybe there's a trick here? Like, are they assuming each year is independent? Because if the probability each year is independent, then it's just 1/20 again. But maybe the question is different? Wait, maybe the wording is tricky. Let me parse it again.  "The probability of snow on 1st January is 1/20. What is the probability that snow will fall on the next 1st January?" So, "next" as in the following year? So next year's January 1st. So if each year's snowfall is independent, then the probability remains 1/20. But maybe there's some hidden assumption here. Wait, sometimes in probability problems, they might be talking about conditional probabilities or something else. But this problem doesn't mention any previous events or dependencies. It just states the probability is 1/20. So if each occurrence is independent, then the probability doesn't change. So next year's probability is also 1/20. 
Wait, but let me think again. Maybe the problem is in another language originally, and "next" is being translated as "the next occurrence" or something? But no, it's about "next 1st January". So like, the next time January 1st comes around, which would be the following year. So in terms of probability, assuming each year is an independent event, the probability is the same each year, right? So if the probability of snow on any given January 1st is 1/20, then regardless of previous years, the next one would also be 1/20.   Alternatively, could it be that the problem is asking for the probability that it will snow next year given that it snowed this year? But the problem doesn't say anything about prior snowfall. It just gives the base rate. So I think the answer is 1/20. Let me confirm.  Another angle: sometimes people confuse "probability of snow on a specific day" with "probability of snow in a year" but here it's specifically on 1st January each year. So if the probability each year on that date is 1/20, then next year's is also 1/20. Unless there's a pattern, but the problem doesn't indicate that. So I think the answer is 1/20. So I just need to write that in a box. Wait, maybe I need to make sure if the problem is in a different context. For example, maybe it's a different place where the next January 1st is in the southern hemisphere? But that's overcomplicating. The problem is likely testing understanding of independent events. So the answer is 1/20. </think>  \boxed{\dfrac{1}{20}}
\end{lstlisting}

\clearpage
A sample Shortest Examples from AceReasoner1.1-SFT Dataset \citep{liu2025acereason} \textbf{($\sim$700 tokens)}:

\textbf{$[$Input$]$}: \begin{lstlisting}[breaklines=true]
A train 18 m long passes a pole in 9 seconds. How long will it take to pass a bridge 36 m long?
\end{lstlisting} 
\textbf{$[$Output$]$}: 
\begin{lstlisting}[breaklines=true]
<think> Okay, so I need to figure out how long it takes for a train that's 18 meters long to pass a bridge that's 36 meters long. The problem already says that it takes 9 seconds for the train to pass a pole. Let me start by understanding what that means.  When the train passes a pole, the entire length of the train has to go by the pole. So, if the train is 18 meters long and it takes 9 seconds to pass the pole, that means the front of the train passes the pole, and then 9 seconds later, the end of the train passes the pole. So, in 9 seconds, the train travels a distance equal to its own length, which is 18 meters.  From this, I can calculate the speed of the train. Speed is distance divided by time. So, speed = 18 meters / 9 seconds. Let me compute that. 18 divided by 9 is 2. So, the train's speed is 2 meters per second. Hmm, that seems right.  Now, the question is about passing a bridge that's 36 meters long. To pass a bridge, the train has to cover the entire length of the bridge plus its own length. Why? Because first the front of the train has to cross the entire bridge, and then the end of the train has to exit the bridge. So, the total distance the train needs to cover is the length of the bridge plus the length of the train. So that would be 36 meters plus 18 meters, which is 54 meters.  So, the train needs to travel 54 meters to completely pass the bridge. Since the speed is 2 meters per second, the time it takes is the total distance divided by speed. Time = 54 meters / 2 meters per second. Let me calculate that. 54 divided by 2 is 27. So, it should take 27 seconds.  Wait, let me double-check to make sure I didn't mix up anything. The first part: passing a pole takes 9 seconds for 18 meters. So speed is 18/9 = 2 m/s. Then, bridge is 36 meters. So, when passing the bridge, the front of the train has to go 36 meters to get to the end of the bridge, and then the entire train has to clear it, which is another 18 meters. So total distance is indeed 36 + 18 = 54 meters. Time is 54 / 2 = 27 seconds. Yeah, that makes sense.  Alternatively, maybe someone might think you just need to cover the bridge's length, but no, the train's own length has to be considered too. Because until the last part of the train is off the bridge, the bridge isn't completely passed. So, adding the two lengths is correct. Therefore, 27 seconds should be the right answer. </think>  The train's speed is \( \frac{18 \text{ m}}{9 \text{ s}} = 2 \text{ m/s} \). To pass a 36 m bridge, the total distance needed is \( 36 \text{ m} + 18 \text{ m} = 54 \text{ m} \). Time required is \( \frac{54 \text{ m}}{2 \text{ m/s}} = 27 \text{ seconds} \).  \boxed{27}
\end{lstlisting}

\clearpage
\subsection{Longest Examples}
A sample Longest Examples from AceReasoner1.1-SFT Dataset \citep{liu2025acereason} \textbf{($\sim$8k tokens)}:

    \textbf{$[$Input$]$}: \begin{lstlisting}[breaklines=true]
    For sets $A$, $B$, and $C$, where $n(A)$, $n(B)$, and $n(C)$ denote the number of subsets of $A$, $B$, and $C$ respectively, and $\left | A \right | = \left | B \right | = 100$, if $n(A) + n(B) + n(C) = n(A \cap B \cap C)$, what is the minimum possible value of $\left | A \cap B \cap C \right |$?} 
    \end{lstlisting}
    
\textbf{$[$Output$]$}: (Full response available at \url{https://huggingface.co/datasets/nvidia/AceReason-1.1-SFT/viewer?views%5B%5D=train&row=0}).

\end{appendices}
\end{document}